# MIGONet: a multiple-input graph neural operator with solver-free discrete manufactured-solution training

Muhammad Hasnain* (محمد حسنين), N. K. Anand (ಎನ್. ಕೆ. ಆನಂದ್), Arun Srinivasa (ಅರುಣ್ ಶ್ರೀನಿವಾಸ)

Department of Mechanical Engineering, Texas A&M University, 3123 TAMU
College Station, TX 77843-3123, USA

*Corresponding Author Email: mhasnain5924117@tamu.edu

**Short title: Graph neural operator with manufactured-solution training**

## Abstract

Surrogate modeling of transport requires operators that capture the combined influence of spatially distributed sources and Dirichlet and Neumann boundary conditions, together with training datasets that can be generated at low computational cost. This work presents MIGONet, a multiple-input graph neural operator, and a discrete method of manufactured solutions (MMS) that trains it without solving a partial differential equation, using steady heat conduction as the model problem. In the discrete MMS procedure, temperature fields are prescribed *a priori*, and the source terms and Dirichlet or Neumann boundary data are evaluated with the same discrete operators as the reference finite-difference method (FDM), so every training sample satisfies the discrete governing equations by construction. Because the boundary data are derived from each field, a Dirichlet or Neumann condition can be assigned to every boundary segment of every sample, and the boundary specification becomes an operator input. MIGONet supplies representations of the complete source field and boundary data to every node before local message passing. On the rectangular domain, it is compared with a version without the complete-source encoder (MIGONet-local) and with the multiple-input operator network MIONet, using the same training data. MIGONet achieves mean relative temperature errors of 10.31% and 10.64% for curved-wire and Gaussian-spot heating, compared with 45.97% and 48.34% for MIONet and 72.25% and 67.82% for MIGONet-local. For 10,000 samples, discrete MMS reduces data-generation time from 486.65 s to 3.15 s relative to matched iterative FDM labeling. Separately trained MIGONet models are also evaluated on L-shaped, disc-shaped, and ring-shaped domains.

## 1 Introduction

Scientific machine learning (SciML) offers data-driven surrogates for repeated solution of transport problems in design exploration, optimization, and uncertainty quantification. For boundary-value problems governed by partial differential equations (PDEs), the solution field is determined by the domain geometry, material properties, internal sources, and boundary conditions. Constructing conventional supervised training datasets can require many numerical reference solutions, making data generation a substantial part of the surrogate-development effort [1]. In addition, predictive performance depends on the coverage of the training data and their consistency with the governing equations and boundary conditions [2], [3], [4]. These considerations motivate data-generation procedures that avoid separately solving the governing equation for every training sample, together with architectures that can represent the influence of spatially distributed sources and mixed boundary conditions.

Neural operators learn mappings between function spaces and are therefore well suited to approximating the solution operators of PDEs [5]. The Deep Operator Network (DeepONet) [6] and the Fourier Neural

Operator (FNO) [7] are two widely used formulations. Many boundary-value problems depend on several input functions. In steady diffusion, the source distribution and prescribed boundary data jointly determine the temperature field and may vary independently between problems. Multiple-input operator networks (MIONet) represent this dependence through separate branch networks for the input functions and a trunk network for the output coordinates [8]. MIONet therefore provides a global multiple-input baseline for evaluating the proposed graph architecture.

Graph- and mesh-based learning methods represent discretized domains through nodes and edges and have been applied to problems involving nonrectangular domains and unstructured meshes [9], [10], [11], [12]. For elliptic problems such as steady diffusion, the solution operator is nonlocal: the temperature at a given location depends on source and boundary information distributed throughout the domain. In a network built from local message-passing layers, the direct propagation of locally supplied source information is limited by the number of graph updates. Increasing the depth also requires information from progressively larger neighborhoods to be compressed into fixed-size node representations [13]. This motivates supplying global source and boundary embeddings to every node before local message passing.

Boundary conditions require explicit treatment because both their type and prescribed values may vary between problems. Dirichlet conditions prescribe temperature, whereas Neumann conditions prescribe outward heat flux, requiring distinct input representations and normalization. Physics-informed and self-supervised operator-learning approaches incorporate governing-equation or boundary-condition residuals to reduce dependence on labeled solution fields [14], [15], [16]. Boundary-aware and label-free formulations also address the representation and enforcement of boundary information [17], [18], [19]. A complementary question is whether a supervised graph model can learn the response to varying sources and mixed boundary specifications from training pairs constructed directly to satisfy the numerical equations.

Solution-first data generation follows the principle of the method of manufactured solutions (MMS), which is widely used for verification of numerical implementations: a solution field is prescribed, and the corresponding source and boundary data are derived from the governing equation [20]. Hasani and Ward [21] sampled candidate solution fields and differentiated them to obtain the corresponding forcing, while Sojitra and San [4] developed a method of manufactured learning using analytically constructed solution families. Dong et al. [22] proposed Differential Operator Action in Solution Space (DiffOAS), which applies differential operators to combinations of basic solution fields to generate additional PDE data. Bischof et al. [15] combined manufactured-solution supervision with physics-informed learning in HyPINO for varying source terms, geometries, and mixed Dirichlet–Neumann conditions.

The study by Choubey et al. [23] is particularly relevant to the present work. They generated random temperature fields using Gaussian-process regression conditioned on prescribed boundary values, evaluated the corresponding sources using finite differences, and trained DeepONet heat-conduction surrogates without forward PDE solutions. Their study also considered nonrectangular domains and temperature-dependent conductivity. In their formulation, the boundary conditions are imposed on the generated fields, whereas the present approach derives both the source and boundary data from each manufactured field. This enables training samples with varying Dirichlet–Neumann assignments and corresponding boundary values.

The present work addresses how discretely consistent manufactured data can be used to train an operator for varying source and boundary configurations while accounting for the nonlocal response of steady diffusion. Two complementary components are developed. First, a discrete MMS procedure generates temperature fields and evaluates the corresponding source terms and segment-wise boundary data using the same discrete operators as the reference finite-difference formulation. Because the boundary data are derived from the manufactured fields, their type and values can vary between segments and samples, while each training pair satisfies the discrete equations by construction. Second, MIGONet is introduced as a multiple-input graph neural operator that combines local node features with separate global representations of the complete source field and boundary specification. These representations are supplied to every node before local message passing, while prescribed Dirichlet temperatures are enforced exactly at the output. This formulation enables a single model to learn from varying source and boundary configurations within each discretized geometry.

The framework is developed and evaluated using steady, isotropic heat conduction with constant conductivity on a fixed $32 \times 32$ Cartesian background grid. Manufactured temperature fields are generated by Gaussian smoothing of random noise with independently sampled smoothing widths in the two coordinate directions. Although the discrete MMS construction can, in principle, be extended to other boundary-value problems using their corresponding discrete governing and boundary operators, only steady heat conduction is implemented and evaluated here.

The rectangular-domain study compares MIGONet with MIGONet-local, which removes the complete-source encoder, and MIONet using common training data, normalization, loss formulation, and Dirichlet enforcement. Generalization is evaluated on boundary-condition cases and independently constructed sinusoidal, curved-wire, Gaussian-spot, and uniform heat-source distributions excluded from training and validation. The same data-generation and learning procedure is then applied to separately trained MIGONet models on L-shaped, disc-shaped, and ring-shaped domains to examine a re-entrant corner, a curved external boundary, and an internal boundary.

The primary contributions of this work are as follows:

i. MIGONet is developed by combining local graph features with separate global embeddings of the complete source field and the segment-resolved boundary specification, both treated as operator inputs. The global embeddings are supplied to every node before local message passing, while prescribed Dirichlet temperatures are enforced exactly at the output.

ii. A discrete manufactured-solution procedure is formulated for PDE boundary-value problems and developed for mixed Dirichlet–Neumann heat conduction. Boundary data are derived from each prescribed temperature field, so a Dirichlet or Neumann condition can be assigned to each boundary segment of every sample. The source, boundary, and temperature data are generated without solving a PDE for each sample and satisfy the reference discrete equations by construction.

iii. The predictive performance of MIGONet is compared with MIGONet-local and MIONet on the rectangular domain using common training data and evaluation cases constructed independently of the MMS training generator. Separately trained MIGONet models assess the applicability of the procedure to L-shaped, disc-shaped, and ring-shaped domains.

iv. The computational cost of training-data construction is quantified by comparing discrete MMS with a matched iterative FDM labeling pipeline.

The remainder of the paper is organized as follows. Section 2 presents the governing equations and discrete formulation, and Section 3 describes the manufactured-solution data-generation procedure. Sections 4 and 5 describe the model architectures, training, and evaluation. Section 6 presents the rectangular-domain results and computational-cost comparison, followed by the additional-geometry studies in Section 7. Section 8 presents the summary and limitations, with additional results and implementation details provided in Appendices A, B, and C.

## 2 Governing equation and discrete formulation

### 2.1 Governing equation and boundary conditions

We consider steady, isotropic diffusion with constant conductivity $k = 1$ in a two-dimensional domain $\Omega$. The temperature $T$ and volumetric source $Q$ satisfy:

$$-k\nabla^2 T = Q \qquad \text{in } \Omega \tag{1}$$

The boundary $\partial\Omega$ is partitioned into Dirichlet and Neumann segments $\Gamma_D$ and $\Gamma_N$, respectively. Dirichlet conditions prescribe the temperature $g_D$, whereas Neumann conditions prescribe the outward heat flux $g_N$:

$$T = g_D \quad \text{on } \Gamma_D, \qquad -k\nabla T \cdot \mathbf{n} = g_N \quad \text{on } \Gamma_N \tag{2}$$

Here, $\mathbf{n}$ is the outward unit normal. Every problem contains at least one Dirichlet boundary segment for well-posedness. When Dirichlet and Neumann assignments meet at a common node, the Dirichlet value takes precedence.

### 2.2 Discrete formulation

All domains are represented on the same $32 \times 32$ Cartesian background grid over the unit square $[0,1] \times [0,1]$, with grid spacing $h = 1/31$. The rectangle uses the complete grid, while the L-shaped, disc-shaped, and ring-shaped domains are represented by fixed active-node masks as shown in Figure 1. At an active interior node $(i,j)$, the discrete Laplacian is the standard five-point operator given by:

$$(\Delta_h T)_{i,j} = \frac{T_{i+1,j} + T_{i-1,j} + T_{i,j+1} + T_{i,j-1} - 4T_{i,j}}{h^2} \tag{3}$$

At an axis-aligned Neumann boundary node $b$, let $a$ denote its adjacent active node in the inward normal direction. Applying a one-sided approximation of the outward normal derivative to the boundary condition in Eq. (2) gives:

$$T_b - T_a = -\frac{h}{k} g_N \tag{4}$$

For a curved boundary with outward normal components $n_x$ and $n_y$, let $a_x$ and $a_y$ denote the adjacent active nodes in the inward $x$- and $y$-directions. Resolving the normal derivative into its Cartesian components and applying one-sided differences from the boundary toward the active domain, the corresponding discrete normal-flux closure can be written as:

$$\left(|n_x| + |n_y|\right)T_b - |n_x| T_{a_x} - |n_y| T_{a_y} = -\frac{h}{k} g_N \tag{5}$$

A directional term is omitted when the corresponding normal component is zero. Thus, Eq. (5) reduces to Eq. (4) when the boundary normal is aligned with a Cartesian axis. The same Cartesian construction is applied to the disc and ring using their local radial normal components. On the inner boundary of the ring, the outward normal of the physical domain points toward the hole. Inactive nodes are assigned zero-valued identity rows in the full background-grid system and are excluded from the physical-domain error metrics.

The manufactured-solution data in Section 3 use this same discrete formulation. A temperature field $T$ is prescribed first, the source $Q$ is evaluated with Eq. (3), and the boundary temperatures or fluxes are obtained from Eqs. (2), (4), and (5). The retained temperature field is therefore consistent with the discrete source and boundary data by construction, without solving a PDE numerically.

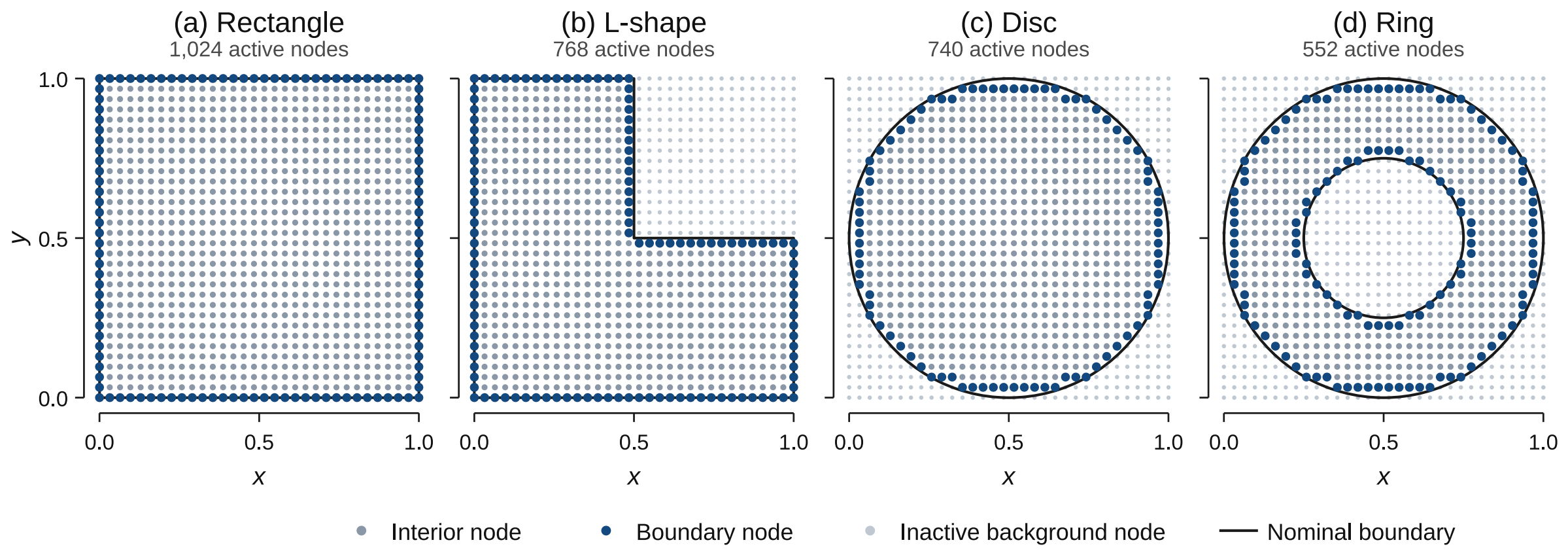

Figure 1. Computational domains represented on the $32 \times 32$ Cartesian background grid: (a) rectangle, (b) L-shape, (c) disc, and (d) ring. Gray and blue markers indicate interior and boundary nodes, faint markers indicate inactive background nodes, and black lines show the nominal domain boundaries. Boundary types vary between samples. A separate MIGONet model is trained for each geometry.

### 2.3 Finite-difference solution procedures

Two types of evaluation cases are used. In the first type, the temperature field is prescribed directly and the corresponding source and boundary data are obtained using the discrete operators defined in Section 2.2. This includes the zero-source boundary-condition tests in Section 6.2 and the sinusoidal-source tests in Section 6.3. Because the temperature field is known by construction, it serves directly as the discrete reference solution and no numerical PDE solve is required. In the second type, the heat-source distribution is prescribed independently, as in the curved-wire, Gaussian-spot, and uniform-heating tests. The corresponding reference temperature field must then be obtained by solving the finite-difference equations assembled from Eqs. (2)–(5). Dirichlet conditions are imposed using identity rows, and inactive nodes in masked geometries are assigned zero-valued identity rows. The sparse linear system $\mathbf{AT} = \mathbf{b}$ is assembled on the complete background grid. These reference solutions are computed using the SuperLU sparse direct LU solver. The direct solution is used for the accuracy studies so that the reported reference fields are not affected by an iterative stopping tolerance.

A separate iterative FDM implementation is used only for the matched data-generation cost comparison in Section 6.6. In this implementation, prescribed Dirichlet temperatures are eliminated from the unknown vector, and the remaining equations are scaled to have unit diagonal entries. All-Dirichlet systems are solved using the conjugate gradient (CG) method, whereas mixed Dirichlet–Neumann systems are solved using the generalized minimal residual (GMRES) method with a restart length of 80. Both iterative solvers use double-precision arithmetic and a relative algebraic residual tolerance of $10^{-6}$.

Writing the assembled finite-difference system as $\mathbf{AT} = \mathbf{b}$, the iterations are terminated when $\frac{\|\mathbf{b}-\mathbf{AT}\|_2}{\|\mathbf{b}\|_2} \leq 10^{-6}$. Thus, the direct and iterative procedures solve the same discrete finite-difference equations but serve different purposes: the direct solver provides the reference fields used for predictive-accuracy evaluation, while the iterative solvers are used for the data-generation timing comparison in Section 6.6.

## 3 Manufactured-solution data generation

The discrete MMS procedure follows from the structure of boundary-value problems. Evaluating the discrete operator and the boundary operators on a prescribed solution is a direct computation, whereas obtaining the solution from given source and boundary data requires solving a linear or nonlinear system. Discrete MMS therefore prescribes the solution, evaluates the source and boundary data from it, and retains the prescribed solution as the target. Because the same discrete operators define the data and the reference solution, each sample satisfies the discrete problem to floating-point precision. In the present work, these operators are the diffusion operator and boundary closures of Section 2.

Figure 2 shows the discrete manufactured-solution data-generation procedure. A temperature field is prescribed first, after which the corresponding source terms and boundary values are computed using the discrete differential operators defined in Section 2. The initial random field ($\xi$) is smoothed using independently sampled correlation lengths, or smoothing widths, $\sigma_x$ and $\sigma_y$. The two correlation lengths are sampled independently from a uniform distribution over $1 - 10$ grid cells.

$$\sigma_x, \sigma_y \sim \mathcal{U}(1,10), \qquad \tilde{T} = G_{\sigma_x,\sigma_y} * \xi, \qquad \xi_{i,j} \sim \mathcal{N}(0,1) \tag{6}$$

Here, $G_{\sigma_x,\sigma_y}$ represents Gaussian smoothing with different correlation lengths along both directions. Larger values of $\sigma_x$ or $\sigma_y$ produce smoother temperature variations along the corresponding direction, while unequal values generate fields with different spatial correlation lengths in x- and y-directions. This anisotropic smoothing broadens the range of spatial structures represented in the training data by allowing different correlation lengths in the two coordinate directions.

Because the amplitude of the smoothed field depends on the sampled correlation lengths and random realization, the field is subsequently rescaled to provide explicit control over the temperature scale. In particular, strongly smoothed fields can exhibit a relatively small temperature range even when their spatial structure remains distinct. Therefore, each smoothed field is scaled so that its maximum absolute temperature, defined here as the reference temperature amplitude $\mathrm{A} = \max_{(\mathrm{i,j}) \in \Omega} |\mathrm{T_{ij}}|$, is sampled uniformly from 15 to 80 °C.

$$\mathrm{A} \sim \mathcal{U}(15,80)\ ^\circ\mathrm{C}, \qquad \mathrm{T} = \mathrm{A}\frac{\tilde{\mathrm{T}}}{\max|\tilde{\mathrm{T}}|} \tag{7}$$

The sampled amplitude interval applies to the reference amplitude $A = \max|T|$, rather than to every nodal temperature value. Thus, sampling $A$ from 15 to 80 °C controls the overall magnitude of the manufactured field without requiring every temperature value to lie within that interval. Because the underlying random field may contain both positive and negative values, the scaled manufactured temperature field may likewise contain values of either sign.

For the positive prescribed-source evaluation cases considered later, the source magnitude is scaled so that the peak reference temperature satisfies $T_{max} = A$. Once the temperature field is defined, the source term is calculated as $Q = -k\Delta_h T$ using the discrete diffusion operator. Each boundary segment is then assigned either a Dirichlet or Neumann condition, with the corresponding boundary temperature or flux obtained from the same temperature field. Because the boundary values are obtained from the manufactured field rather than imposed on it, no conditioning or boundary-fitting step is required. The resulting source field, boundary data, and target temperature field are therefore consistent with the discrete formulation by construction, and no numerical PDE solution is required to generate the training data.

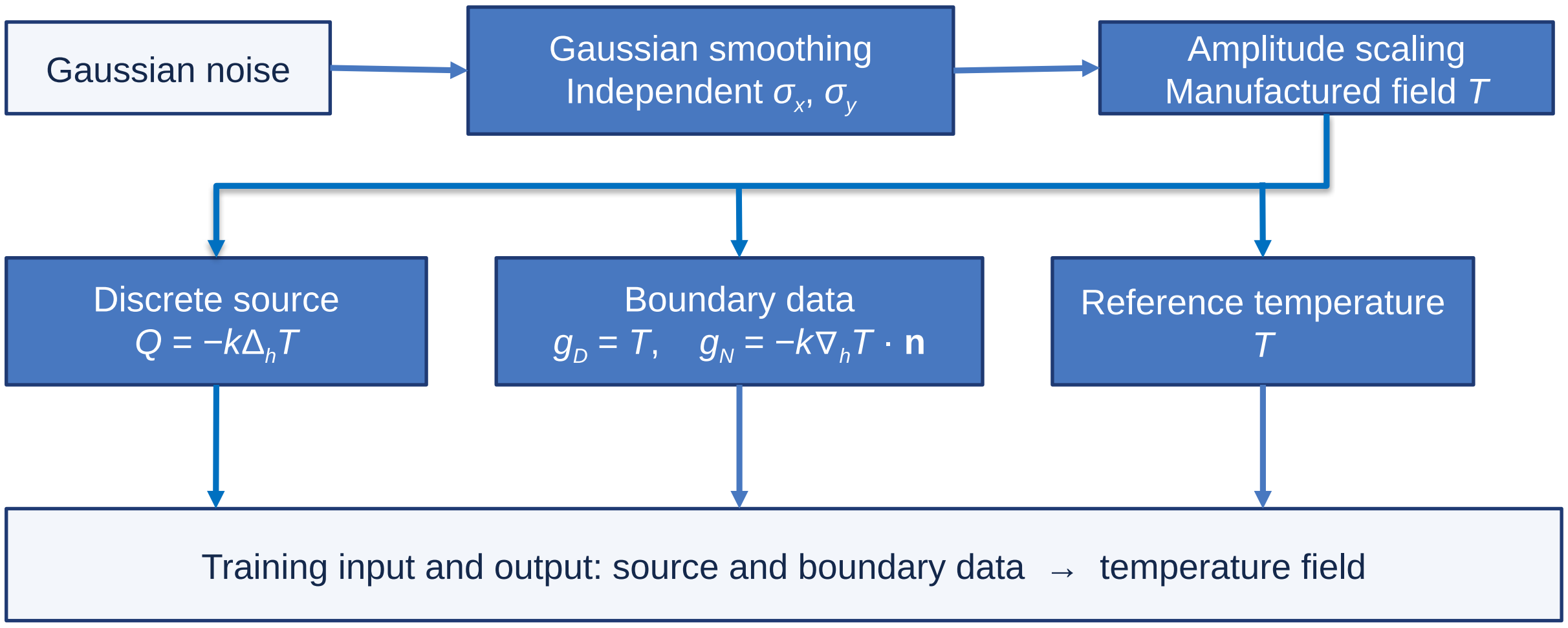

Figure 2. Discrete MMS data generation. A temperature field $T$ is prescribed first. The source $Q$ and boundary data are calculated from that field, which is retained as the training target.

Each geometry uses 10,000 training samples and 1,000 validation samples generated using the discrete MMS procedure described above. Model performance is evaluated on unseen zero-source boundary-condition cases and prescribed-source cases, including curved wires, Gaussian spots, uniform heating, and the sinusoidal-source tests considered in Section 6. These evaluation cases are excluded from both the training and validation datasets. The computational benefit of MMS-based training-data generation relative to conventional FDM-based label generation is quantified separately in Section 6.6.

The source and temperature fields are standardized independently using z-score normalization based on the training-set statistics:

$$Q^* = \frac{Q-\mu_Q}{s_Q}, \qquad T^* = \frac{T-\mu_T}{s_T} \tag{8}$$

Here, $\mu_Q$ and $s_Q$ denote the mean and standard deviation of the source field, respectively, $\mu_T$ and $s_T$ denote the corresponding statistics of the temperature field, and $Q^*$ and $T^*$ represent the standardized source and temperature fields. This scaling keeps the network inputs and targets on comparable numerical scales during optimization. All normalization statistics are computed from the training set and reused unchanged for validation and evaluation.

Because Dirichlet boundary values represent prescribed temperatures, they are normalized using the temperature statistics. Neumann boundary values represent outward heat flux and are therefore normalized using separate flux statistics:

$$b_D^* = \frac{g_D-\mu_T}{s_T}, \qquad b_N^* = \frac{g_N-\mu_N}{s_N} \tag{9}$$

Here, $\mu_N$ and $s_N$ are the mean and standard deviation of the Neumann boundary fluxes in the training data. All normalization statistics are computed from the training set only and are subsequently reused during validation and evaluation.

Each node is additionally assigned a type indicator ( $\tau$), with $\tau = 0$ for interior nodes, $\tau = 1$ for Neumann boundary nodes, and $\tau = 2$ for Dirichlet boundary nodes. The boundary-value feature is set to zero at interior nodes. For the nonrectangular geometries, temperature-amplitude scaling and temperature normalization statistics are evaluated over the active nodes only, whereas source statistics are computed over interior nodes. Inactive background nodes are excluded from these calculations. Because every sample contains at least one Dirichlet segment, solving the discrete system with the manufactured source and boundary data returns the manufactured field. The MMS and FDM labels therefore coincide for identical inputs, and the two procedures differ only in how the inputs are sampled and in the cost of obtaining the target.

# 4 Model architectures

MIGONet maps the source field and boundary-condition specification to the corresponding nodal temperature field. The discretized domain is represented as a graph in which each active grid point is a node and horizontal and vertical nearest neighbors are connected by edges. For the rectangular domain, all ($32 \times 32 = 1024$) grid points are active. The architecture combines three sources of information: local node features, the complete boundary specification, and the complete source field. The local features vary from node to node, whereas the boundary and source encodings represent the complete input configuration and are supplied to every node before message passing. Figure 3 summarizes this architecture and shows the local and global inputs to the architecture.

## 4.1 Local and global input encoding

For node $u$, the local input vector is given by:

$$\mathbf{z}_u = (x_u, y_u, Q_u^*, \tau_u, b_u^*) \tag{10}$$

where $(x_u, y_u)$ are the node coordinates, $Q_u^*$ is the normalized local source value, $\tau_u$ is the node-type indicator defined in Section 3, and $b_u^*$ is the normalized boundary-value feature. As defined previously, $b_u^* = 0$ at interior nodes, while boundary nodes contain the corresponding normalized Dirichlet temperature or Neumann flux.

The five-dimensional local input is mapped to a 128-dimensional latent representation by the local encoder. In addition to this nodewise information, MIGONet uses separate encoders for the complete boundary specification and source field. For the rectangular domain, the four boundary segments contain 128 boundary entries. The associated node types and normalized boundary values are concatenated to form the boundary vector $\mathbf{c} \in \mathbb{R}^{256}$. The boundary encoder therefore receives 128 node-type entries and 128 normalized boundary values. The source encoder receives the complete normalized $32 \times 32$ source field, flattened to a 1024-dimensional vector.

The three encoders produce 128-dimensional representations. The boundary and source embeddings are computed once for each sample and broadcast to every node. The initial latent representation at node $u$ is then obtained by adding the outputs of all three encoders:

$$\mathbf{h}_u^{(0)} = E_{\text{local}}(\mathbf{z}_u) + E_{\text{BC}}(\mathbf{c}) + E_Q\big(\text{vec}(Q^*)\big) \tag{11}$$

where $E_{\text{local}}$, $E_{\text{BC}}$, and $E_Q$ denote the local, boundary, and complete-source encoders, respectively. The local encoder has widths $5 \to 128 \to 128$, the boundary encoder has widths $256 \to 128 \to 128$, and the source encoder has widths $1024 \to 128 \to 128$. Gaussian error linear unit (GELU) activation is applied between the linear layers.

This construction gives every node direct access to the complete boundary and source configuration before graph propagation begins, while retaining the node-specific source, location, type, and boundary information in the local representation. The three inputs correspond to the source, boundary specification, and node layout that define a steady boundary-value problem. The diffusion operator enters the learning problem only through the training data.

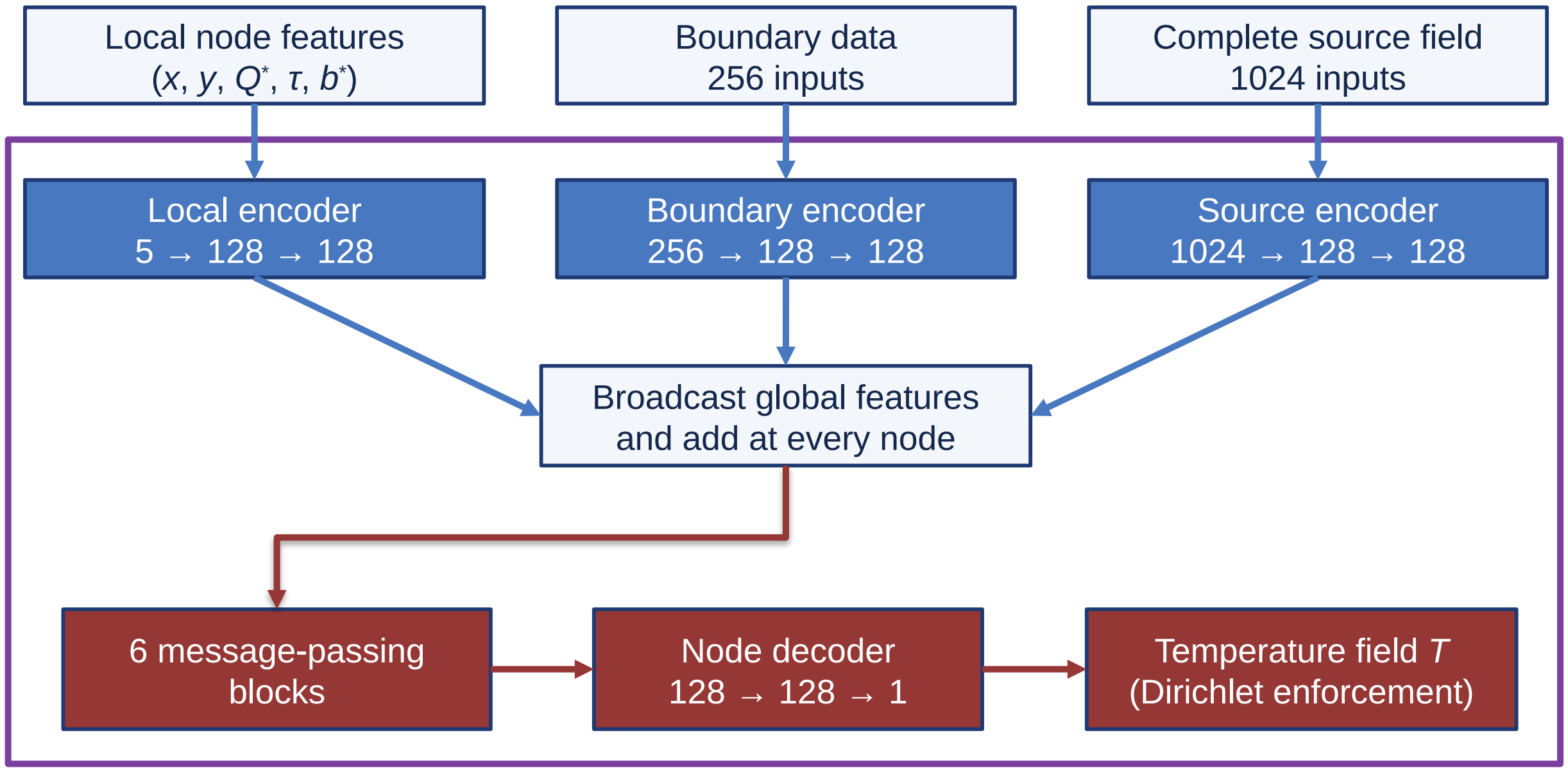

Figure 3. MIGONet architecture for the rectangular domain. Nodewise features are combined with global boundary and complete-source embeddings before six message-passing blocks. A nodewise decoder predicts the normalized temperature, followed by exact enforcement of Dirichlet boundary values. MIGONet-local removes the complete-source branch that encodes the global source field.

### 4.2 Message passing and temperature prediction

After input encoding, six message-passing layers propagate information through the graph. At layer $\ell$, node $u$ aggregates transformed latent features from its neighboring nodes $v \in \mathcal{N}(u)$:

$$\mathbf{m}_u^{(\ell)} = \sum_{v\in\mathcal{N}(u)} \left(W_m^{(\ell)}\mathbf{h}_v^{(\ell)} + \mathbf{b}_m^{(\ell)}\right) \tag{12}$$

where $\mathcal{N}(u)$ denotes the set of graph neighbors of node $u$, and $W_m^{(\ell)}$ and $\mathbf{b}_m^{(\ell)}$ are trainable parameters of the $l$-th message-passing layer.

The aggregated message $\mathbf{m}_u^{(\ell)}$ is combined with the current node representation and passed through an update multilayer perceptron (MLP). A residual connection and layer normalization are then applied:

$$\mathbf{h}_u^{(\ell+1)} = \mathrm{LayerNorm}\left[\mathbf{h}_u^{(\ell)} + F^{(\ell)}\left(\left[\mathbf{h}_u^{(\ell)}; \mathbf{m}_u^{(\ell)}\right]\right)\right] \tag{13}$$

The update MLP, $F^{(\ell)}$, is configured with widths of $256 \rightarrow 128 \rightarrow 128$ with GELU activation between the linear layers. Because each graph layer communicates between directly connected nodes, six message-passing layers allow locally supplied information to propagate over paths of up to six graph edges. The global boundary and source representations, however, are available to every node before these local updates begin.

After the sixth graph layer, a decoder with widths $128 \rightarrow 128 \rightarrow 1$ maps each latent node representation to a normalized temperature prediction. Hard constraints are applied for Dirichlet boundary nodes by replacing the network output with the prescribed normalized temperatures during both training and evaluation:

$$\hat{T}_u^* = \begin{cases} b_u^* \;, & (\tau_u = 2) \\ D\left(\mathbf{h}_u^{(6)}\right) \;, & (\tau_u \neq 2) \end{cases} \tag{14}$$

where $D$ denotes the node decoder. The corresponding physical temperature is subsequently recovered using $\hat{T} = s_T \hat{T}^* + \mu_T$.

Thus, Dirichlet conditions are satisfied exactly by construction. Neumann conditions are supplied to the network through the boundary features and global boundary encoding and influence the predicted field through the learned mapping; they are not imposed through output replacement.

### 4.3 Comparison architectures: MIGONet-local and MIONet

Two comparison architectures are considered to evaluate the role of complete-source information and to compare the proposed graph-based operator with a multiple-input neural operator.

MIGONet-local retains the same local encoder, global boundary encoder, message-passing layers, decoder, and Dirichlet enforcement as MIGONet, but removes the global-source encoder $E_Q$. The local source value $Q_u^*$ remains part of the node feature vector in Eq. (10). Consequently, source information originating at one node can influence another node only through successive message-passing updates. With six message-passing layers, locally supplied source information can propagate over at most six graph edges. In contrast, MIGONet encodes the complete source field and broadcasts its latent representation to every node before message passing. This distinction is important for diffusion problems because the temperature at a given location generally depends on heat generation over a region much larger than the local message-passing neighborhood.

The second comparison architecture is MIONet [8], an operator-learning architecture developed for mappings with multiple functional inputs. MIONet provides a particularly relevant comparison because it processes the complete source field and boundary data through separate global branches and uses the spatial coordinates as the trunk input, as shown in Figure 4, whereas MIGONet combines global representations of these inputs with graph-based local interactions.

To maintain consistent physical inputs across the comparison, MIONet receives the same normalized complete source field, boundary representation, and spatial coordinates used by MIGONet. The boundary branch has widths 256 → 512 → 1024 → 1000; the source branch has widths 1024 → 1024 → 1024 → 1000; and the trunk has widths 2 → 128 → 256 → 512 → 1024 → 1000. GELU activation is used between linear layers. The boundary and source branches each produce a 1000-dimensional latent representation, followed by layer normalization. Their outputs are combined through an elementwise product and then contracted with the coordinate-trunk representation through a weighted dot product:

$$\hat{T}^*(\mathbf{x}) = \sum_{r=1}^{1000} w_r \, b_{\mathrm{BC},r}(\mathbf{c}) \, b_{Q,r}(Q^*) \, t_r(\mathbf{x}) + b_0 \tag{15}$$

where $b_{\mathrm{BC},r}$, $b_{Q,r}$, and $t_r$ denote the $r$-th components of the boundary branch, source branch, and coordinate trunk, respectively, $w_r$ are trainable combination weights, and $b_0$ is a scalar bias.

After evaluating the MIONet output at each node, the predicted values at Dirichlet nodes are replaced by the corresponding prescribed boundary temperatures. This enforces the Dirichlet conditions exactly and ensures that the same hard boundary enforcement is applied to all three models.

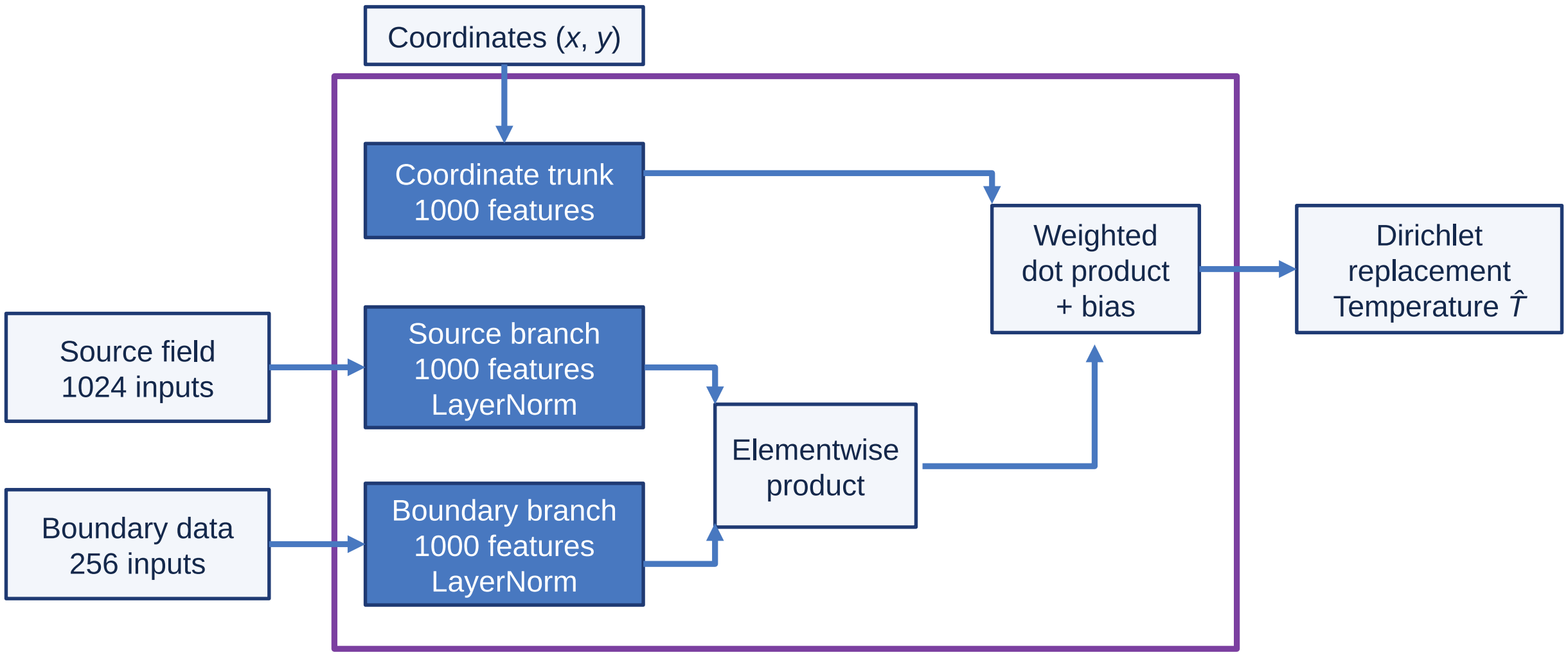

Figure 4. MIONet architecture used for comparison. The complete source field and boundary data are processed by separate branch networks, whose latent representations are combined by an elementwise product and contracted with the coordinate-trunk features to obtain the nodal temperature prediction. Dirichlet boundary values are subsequently enforced exactly.

The comparison is particularly relevant for steady diffusion because the underlying elliptic solution operator is nonlocal: the temperature at a given location can depend on source and boundary information distributed throughout the domain. For a fixed boundary-condition configuration, this global dependence can be expressed formally through Green's-function representations involving contributions from the interior source field and boundary data. This motivates architectures that provide access to the complete problem specification rather than relying exclusively on local spatial propagation. MIONet provides a fully global operator-learning baseline by processing the complete source and boundary inputs through global branches, whereas MIGONet combines global source and boundary embeddings with graph-based local interactions. MIGONet-local removes the complete-source pathway and therefore provides a direct test of whether local message passing alone is sufficient to represent the long-range influence of distributed heat generation.

Taken together, the three architectures provide two complementary comparisons. The MIGONet–MIGONet-local comparison evaluates the benefit of supplying complete-source information to a graph model beyond the information available through local message passing. The MIGONet–MIONet comparison evaluates whether a graph-based operator that combines local connectivity with global physical inputs can achieve competitive predictive accuracy relative to a fully global multiple-input operator while using substantially fewer trainable parameters. Table 1 summarizes the source representation and trainable parameter count for each model.

Table 1. Source representation and trainable parameter counts for the rectangular-domain models.

| **Model** | **Source representation** | **Trainable parameters** |
|---|---|---|
| MIGONet-local | Local source values | 480,385 |
| MIONet | Complete source field | 6,526,401 |
| MIGONet | Local and complete source fields | 628,097 |

The global source encoder adds 147,712 trainable parameters relative to MIGONet-local. MIGONet nevertheless contains only approximately 9.6% of the trainable parameters of MIONet. Because the MIGONet–MIGONet-local comparison changes both the available source information and the model capacity, it quantifies the practical effect of introducing the complete-source pathway rather than a strictly parameter-matched architectural ablation.

# 5 Training and evaluation

## 5.1 Training objective and optimization

All models are trained using the normalized manufactured-solution data described in Section 3. The training objective combines a nodal temperature error with an interior gradient-matching term so that the optimization accounts for both field values and local spatial gradients. For $e^* = \hat{T}^* - T^*$, the objective is

$$\mathcal{L} = \mathrm{mean}[(e^*)^2] + \lambda_g \mathcal{L}_g \tag{16}$$

where $\lambda_g$ controls the contribution of the gradient term, $\mathcal{L}_g$, defined as:

$$\mathcal{L}_g = \left(\frac{s_T}{h g_{\mathrm{rms}}}\right)^2 \left\{\mathrm{mean}[(\delta_x e^*)^2] + \mathrm{mean}\left[\left(\delta_y e^*\right)^2\right]\right\} \tag{17}$$

In Eq. (17), $s_T$ is the training-set temperature standard deviation used for normalization, $h$ is the grid spacing, $g_{\mathrm{rms}}$ is the root-mean-square temperature-gradient magnitude of the training targets, and $\delta_x$ and $\delta_y$ denote first differences between adjacent interior nodes. The factor $\frac{s_T}{h}$ converts differences in normalized temperature to the physical temperature-gradient scale, while normalization by $g_{\mathrm{rms}}$ keeps the gradient contribution relative to the characteristic gradients in the training data. The value-loss term is evaluated over the complete physical field after exact Dirichlet enforcement. The gradient term compares predicted and reference temperature gradients; it is not a PDE-residual loss and does not introduce additional solver-based supervision.

Optimization settings are summarized in Table 2. The same training procedure is used for MIGONet, MIGONet-local, and MIONet on the rectangular domain so that the comparison is based on common data, loss formulation, normalization, and boundary enforcement. The additional-geometry MIGONet models are trained separately on their geometry-specific MMS datasets using the same optimization settings.

Table 2. Training settings used for all models.

| **Setting** | **Value** |
|---|---|
| Optimizer / batch size | AdamW / 32 |
| Initial learning rate / weight decay | $3 \times 10^{-4}$ / $10^{-5}$ |
| Gradient weight / clipping norm | $\lambda_g = 0.1$ / 30 |
| Maximum epochs / early-stop patience | 500 / 75 epochs |
| Learning-rate reduction | Factor 0.5 after 20 plateau epochs |
| Checkpoint selection | Lowest validation mean squared error (MSE) |
| Activation | GELU |

The checkpoint selected by the validation-temperature MSE criterion in Table 2 is retained for evaluation. No evaluation case is used for parameter fitting, learning-rate scheduling, early stopping, or checkpoint selection.

### 5.2 Evaluation metrics and generalization protocol

The primary accuracy measure is the relative discrete $L_2$ temperature error:

$$\varepsilon_{L_2} = 100\frac{\|\hat{T}-T_{\text{ref}}\|_2}{\|T_{\text{ref}}\|_2} \tag{18}$$

where $\hat{T}$ is the predicted temperature field, $T_{\text{ref}}$ is the corresponding reference field, and $\|\cdot\|_2$ denotes the discrete Euclidean norm. For the rectangular domain, the norm is evaluated over the complete physical grid; for the nonrectangular domains, only active nodes are included. For each source family, the reported mean error is obtained by averaging the case-level errors over the corresponding source configurations.

Contour comparisons in the results section also report the absolute nodal error:

$$e_{\text{abs}}(x,y) = \left|\hat{T}(x,y) - T_{\text{ref}}(x,y)\right| \tag{19}$$

The evaluation benchmarks are constructed independently of the random MMS training generator described in Section 3. Boundary cases alter the wall-condition configuration, while the prescribed-source cases use structured source distributions that are not sampled directly by the random-field generator. The evaluation tests generalization to unseen prescribed source distributions and boundary-condition combinations while the reference amplitudes ($A$ = 15, 35, and 55 °C) remain within the trained amplitude range. At each amplitude, the rectangular comparison includes five zero-source boundary-condition cases, two sinusoidal source cases, 36 wire cases, 12 Gaussian-spot cases, and one uniform-source case, giving 56 cases per amplitude and 168 cases per model in total.

## 6 Rectangular-domain results and discussion

The rectangular domain is used for the detailed comparison of MIGONet, MIGONet-local, and MIONet because all three models share the same grid, training data, normalization, loss formulation, and boundary enforcement. The results are organized to separate the effects of the manufactured-field generator, boundary-condition variation, structured source distributions, localized heat generation, source magnitude and width, and training-data generation cost. All predictive cases are excluded from the training and validation datasets, as described in Section 5.

### 6.1 Visualization of manufactured temperature fields

Before evaluating model generalization, Figure 5 illustrates how the two smoothing lengths control the spatial structure of the manufactured temperature fields. The same initial random field and reference amplitude $A = 35\ ^{\circ}\text{C}$ are used in all nine fields so that the effect of correlation lengths or smoothing widths $\sigma_x$ and $\sigma_y$ can be clearly noticed. Small smoothing lengths retain short-scale temperature variations, whereas increasing $\sigma_x$ or $\sigma_y$ produces progressively smoother variation along the corresponding coordinate direction. Unequal smoothing lengths generate different correlation scales in the two directions. These examples are selected to isolate the role of the correlation lengths; during training, the initial random fields, smoothing lengths, amplitude, and boundary assignments all vary independently.

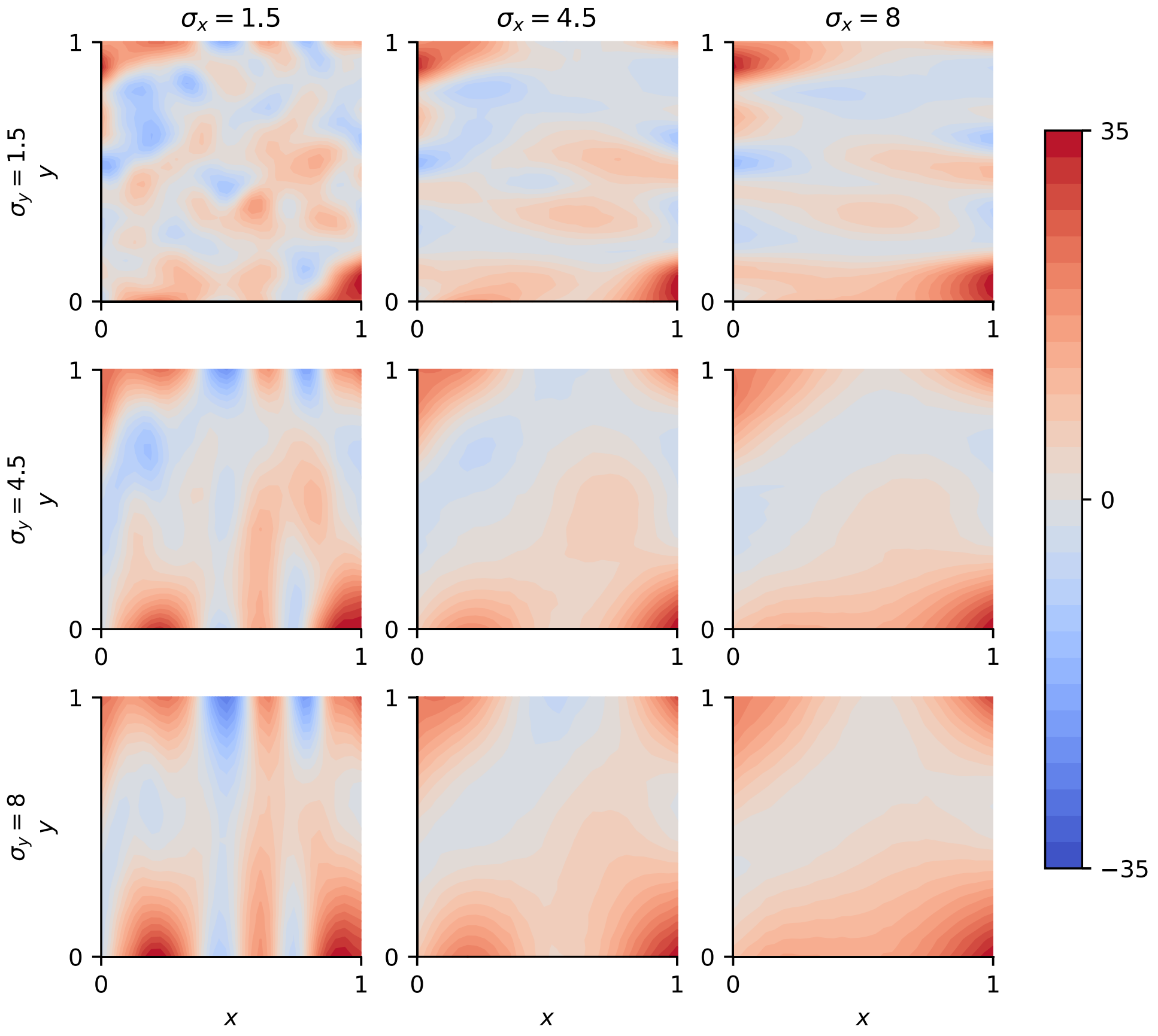

Figure 5. Effect of Gaussian smoothing lengths on manufactured fields. Columns vary $\sigma_x$ and rows vary $\sigma_y$, in grid-cell units. A common noise realization and amplitude $A = 35\ ^\circ C$ isolate the effect of smoothing. These are controlled illustrations of the generator rather than selected training records.

### 6.2 Generalization to unseen boundary conditions

To isolate generalization across boundary conditions, these tests set the internal heat source to $Q = 0$ and vary only the prescribed boundary temperatures and/or heat fluxes. The five configurations in Table 3 span changes in heat-flow direction, Dirichlet and Neumann boundary types, insulation, and applied wall flux. Cases B1 and B2 use all-Dirichlet conditions to produce diagonal and vertical conduction, respectively. B3 retains the same vertical reference field as B2 but replaces the side-wall temperatures with insulation (homogeneous Neumann or no-flux boundary condition), providing a direct change in boundary type without changing the intended temperature distribution.

Table 3. Boundary-condition test cases. D denotes prescribed temperature and N denotes prescribed outward flux. The parameter $A$ is the maximum absolute reference temperature in °C.

| **Case** | **Boundary conditions** | **Reference temperature** |
|---|---|---|
| B1 Diagonal heat flow | All D; wall temperatures vary linearly | $T = A(y - x)$ |
| B2 Vertical heat flow | All D; hot top and cold bottom; linear side temperatures | $T = A(2y - 1)$ |
| B3 Insulated side walls | Top and bottom D; left and right insulated | $T = A(2y - 1)$ |
| B4 Applied wall flux | Left and right D; top and bottom N | $T = A(7x + 13y)/20$ |
| B5 Curved contours | All D; wall temperatures vary quadratically | $T = A(x^2 - y^2)$ |

In B1, the upper-left corner is hottest and the lower-right corner is coldest, producing diagonal heat flow. In B2, $T = A$ is imposed at the top and $T = -A$ at the bottom, with linearly varying side-wall temperatures. B3 uses the same linear reference field but imposes insulated side walls, $g_N = 0$. The B2-B3 pair therefore tests whether the models respond correctly when the boundary specification changes while the reference field is retained.

B4 introduces nonzero Neumann data and is included to test mixed non-homogeneous Dirichlet-Neumann conditions with applied wall fluxes. The top and bottom fluxes are $g_N = -13A/20$ and $g_N = 13A/20$, respectively. B5 uses quadratic wall temperatures to generate curved temperature contours while satisfying $\nabla^2 T = 0$, so no internal source is required. Together, the five cases cover changes in heat-flow direction, insulation, applied flux, and boundary-induced curvature.

Table 4. Relative $L_2$ temperature errors (%) at $A = 35\ ^{\circ}C$.

| **Case** | **MIGONet-local** | **MIONet** | **MIGONet** |
|---|---|---|---|
| B1 Diagonal heat flow | 12.18 | 7.28 | 7.63 |
| B2 Vertical heat flow | 23.73 | 9.64 | 5.75 |
| B3 Insulated side walls | 14.94 | 9.39 | 10.11 |
| B4 Applied wall flux | 36.35 | 39.70 | 11.13 |
| B5 Curved contours | 24.41 | 6.62 | 8.23 |

The relative errors for the five cases are reported in Table 4. MIGONet gives the lowest mean error, 8.57%, compared with 22.32% for MIGONet-local and 14.52% for MIONet. Its largest advantage occurs for B4, where the applied-flux case gives errors of 11.13%, 36.35%, and 39.70% for MIGONet, MIGONet-local, and MIONet, respectively. MIGONet also gives the lowest error for B2, whereas MIONet is slightly more accurate for B1, B3, and B5. Because $Q = 0$ in all five cases, this comparison evaluates the overall treatment of boundary information rather than isolating the effect of complete-source encoding.

Figure 6 shows B3 as a representative mixed-boundary case. The reference temperature varies linearly from the cold bottom wall to the hot top wall and remains constant along each horizontal line. All three models satisfy the top and bottom Dirichlet values exactly through the hard boundary enforcement described in Section 4. The remaining error therefore reflects the interior prediction and the response to the insulated side walls; the horizontal variations visible in the predicted contours indicate departures from the one-dimensional reference field.

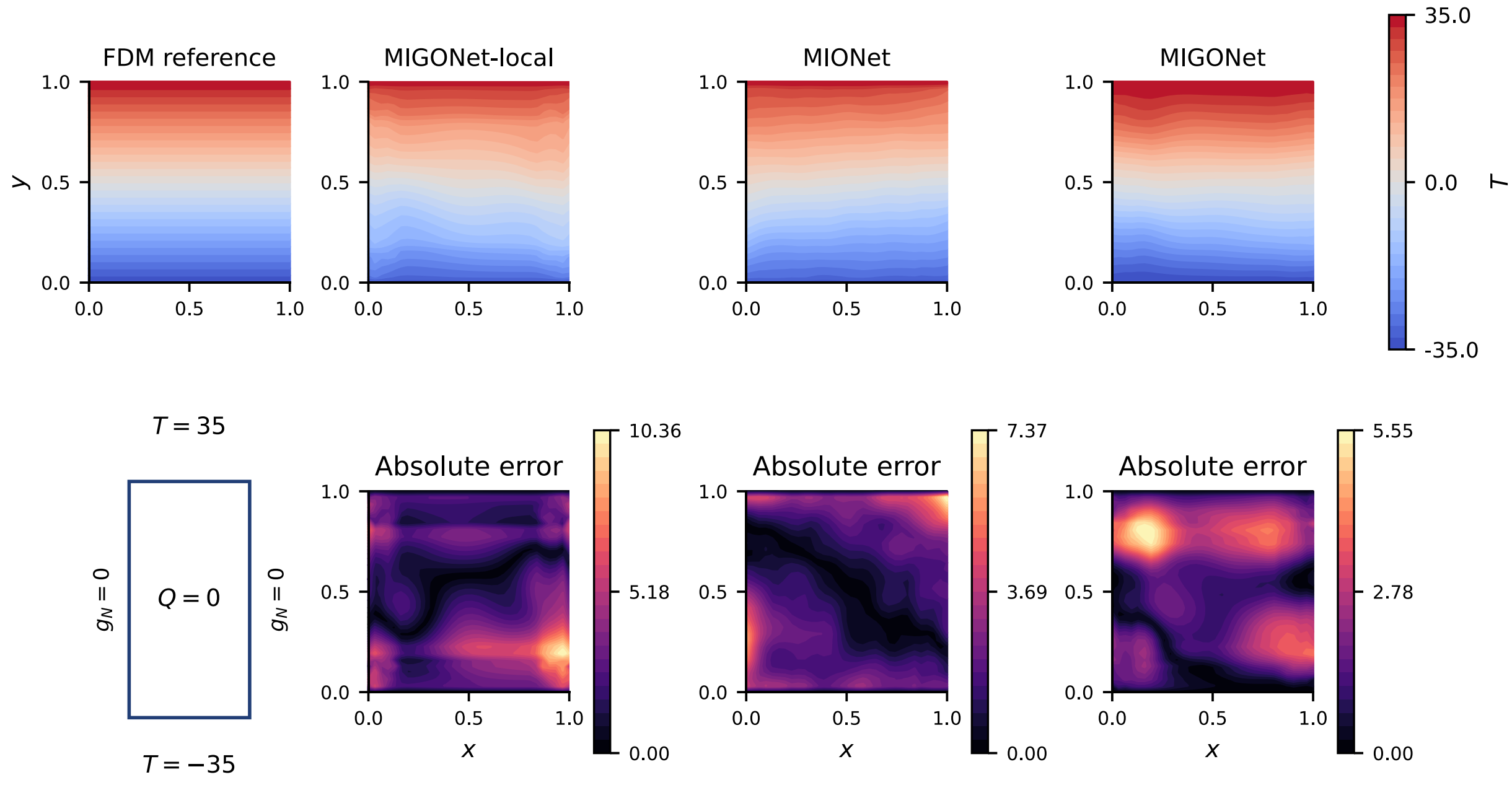

Figure 6. Zero-source conduction with insulated side walls, case B3, at $A = 35\,^{\circ}C$. The upper row compares temperatures. The lower row shows the boundary conditions and absolute errors.

### 6.3 Generalization to sinusoidal source distributions

The sinusoidal-source tests introduce a structured source pattern with alternating heating and cooling regions that is qualitatively different from the randomly smoothed MMS fields used for training. The manufactured temperature shape is defined by Eq. (20) and scaled to the reference amplitude in Eq. (21); the corresponding source is then obtained with the same discrete diffusion operator used in Section 3.

$$S(x, y) = \sin(2\pi x)\cos(3\pi y) + 0.55\sin(4\pi x)\sin(2\pi y) \tag{20}$$

$$T = A\frac{S}{\max|S|}, \qquad Q = -k\Delta_h T \tag{21}$$

Two boundary configurations are evaluated for the same structured temperature and source fields. The first uses Dirichlet conditions on all four walls, whereas the second uses Dirichlet conditions on the left and right boundaries and Neumann conditions on the top and bottom. This pair is selected to separate the effect of boundary specification from the spatial structure of the source. Because the temperature field is manufactured using the same discrete operators as the governing formulation, it serves directly as the discrete reference solution for both boundary configurations; no numerical FDM solve is required for these cases.

Table 5. Relative $L_2$ temperature errors (%) for sinusoidal sources at $A = 35\,^{\circ}C$.

| **Boundary conditions** | **MIGONet-local** | **MIONet** | **MIGONet** |
|---|---|---|---|
| All walls Dirichlet | 12.99 | 48.19 | 13.21 |
| Mixed Dirichlet–Neumann | 16.43 | 60.58 | 17.97 |

Table 5 shows that both graph models are more accurate than MIONet for the two sinusoidal-source tests. For the all-Dirichlet case, the relative errors are 12.99% for MIGONet-local, 48.19% for MIONet, and 13.21% for MIGONet. For the mixed-boundary case, the corresponding errors are 16.43%, 60.58%, and 17.97%. For this structured field, MIGONet-local is slightly more accurate than MIGONet, by 0.22 and 1.54 percentage points in the two cases. A likely reason is that each term in Eq. (20) is an eigenfunction of the interior five-point operator: for a term with wavenumbers $a$ and $b$, $-\Delta_h$ returns the same term multiplied by the discrete eigenvalue $\lambda_h = (4/h^2)[\sin^2(ah/2) + \sin^2(bh/2)]$. For h = 1/31, the discrete eigenvalues of the two terms are 127.49 and 195.11, which are 0.64% and 1.16% smaller than the corresponding continuous eigenvalues, $13\pi^2$ and $20\pi^2$. The source is, therefore, closely related to $T$ at each interior node, and the local source value in the node features already carries much of the information needed to predict the nodal temperature. These terms are, however, eigenfunctions of the interior operator rather than of the boundary-value problem, because the first term does not vanish at y = 0 and y = 1. Figure 7 shows the all-Dirichlet case and confirms that both graph models recover the principal positive and negative temperature regions more closely than MIONet.

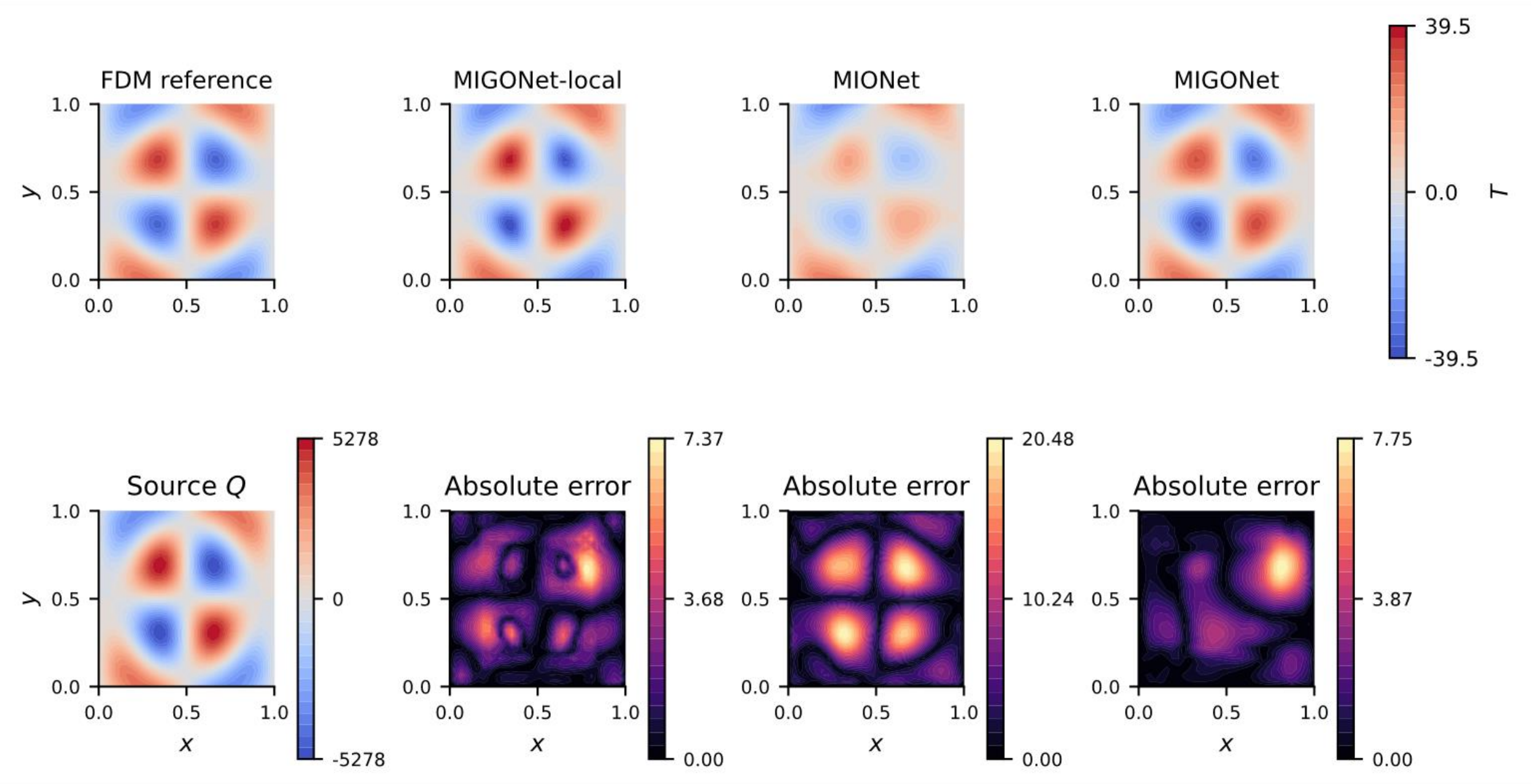

Figure 7. Sinusoidal-source case with all-Dirichlet boundaries at $A = 35\,^{\circ}C$. The lower row contains the source $Q$ and absolute errors $e_{\text{abs}}$.

## 6.4 Generalization to localized and distributed heat-source distributions

These tests prescribe the heat-source distribution independently while imposing homogeneous Dirichlet temperature on all boundaries. Three source families are considered to span different levels of spatial localization: finite-width curved line-like sources, referred to as curved wires; localized Gaussian spots; and uniform heating. The curved wires and Gaussian spots represent line-localized and compact localized heating, respectively, while remaining spatially resolved on the finite-difference grid; uniform heating represents a distributed source over the entire domain. These source distributions differ substantially from the randomly smoothed manufactured fields used for training and therefore provide a more challenging evaluation of source-term generalization. A curved wire is represented by a smooth finite-width source concentrated around an internal centerline. For a spatial location $\mathbf{x} = (x, y)$, the wire source is defined as:

$$Q_\sigma(\mathbf{x}) = \frac{\alpha}{\sqrt{2\pi}\sigma} \int_{\Gamma_w} \exp\left[-\frac{\|\mathbf{x}-\mathbf{r}(s)\|^2}{2\sigma^2}\right] \mathrm{d}\ell \tag{22}$$

Here, $\mathbf{x} = (x, y)$ denotes the spatial location at which the source is evaluated, $\mathbf{r}(s)$ denotes the position of a point along the wire centerline $\Gamma_w$, and $\mathrm{d}\ell$ is the differential arclength along the centerline. The parameter $\sigma$ controls the effective wire width, while $\alpha$ controls the source magnitude. The line integral accumulates the Gaussian contribution from the entire centerline, producing a smooth source field around the curved wire. Before rotation, the centerline is parameterized by Eq. (23).

$$x(s) = 0.2 + 0.6s, \qquad y(s) = c + a\sin(2\pi s + \phi), \qquad 0 \le s \le 1 \tag{23}$$

Thus, the vertical position $c$, curvature amplitude $a$, and phase $\phi$ enter the source distribution through the centerline position $\mathbf{r}(s)$. The centerline is also rotated about the domain center by an orientation angle $\theta$. Twelve independently generated centerlines are used, and each is evaluated at $\sigma = 0.035$, $0.055$, and 0.09, giving 36 wire configurations per amplitude. Detailed parameter ranges, numerical evaluation of the line integral, source-scaling procedure, and other implementation details are provided in Appendix B. Twelve Gaussian spots with varying locations and widths and one uniform-source case provide complementary tests of compact and distributed heating.

All prescribed-source cases use homogeneous Dirichlet conditions, $T = 0$, on every wall. For each independently defined source distribution, the corresponding reference temperature field is obtained by solving the finite-difference system described in Section 2.3 using the SuperLU sparse direct LU solver. The source strength is then scaled so that the peak reference temperature equals $A$. The target temperature field is therefore not specified *a priori* for these cases, and the network is not provided with the target peak. This evaluation directly tests whether a model trained only on random MMS fields can map independently defined source distributions to the corresponding diffusion response.

Table 6. Mean relative $L_2$ errors (%) at $A = 35\ ^{\circ}C$, averaged over 36 wires, 12 spots, or one uniform-source case.

| **Source family** | **MIGONet-local** | **MIONet** | **MIGONet** |
|---|---|---|---|
| Curved wires | 72.25 | 45.97 | 10.31 |
| Gaussian spots | 67.82 | 48.34 | 10.64 |
| Uniform source | 72.05 | 62.73 | 13.08 |

The results in Table 6 show the strongest separation among the three architectures. At $A = 35\ ^{\circ}\mathrm{C}$, MIGONet gives a mean wire error of 10.31%, compared with 72.25% for MIGONet-local and 45.97% for MIONet. For the Gaussian spots, the corresponding errors are 10.64%, 67.82%, and 48.34%. MIGONet also gives the lowest error for the uniform-source case, at 13.08%. The improvement therefore extends across localized line sources, compact spots, and distributed heating rather than being tied to one source geometry.

These errors are obtained for source families that are excluded from training and are not sampled by the random-field generator; they therefore measure out-of-distribution generalization rather than interpolation within the training distribution.

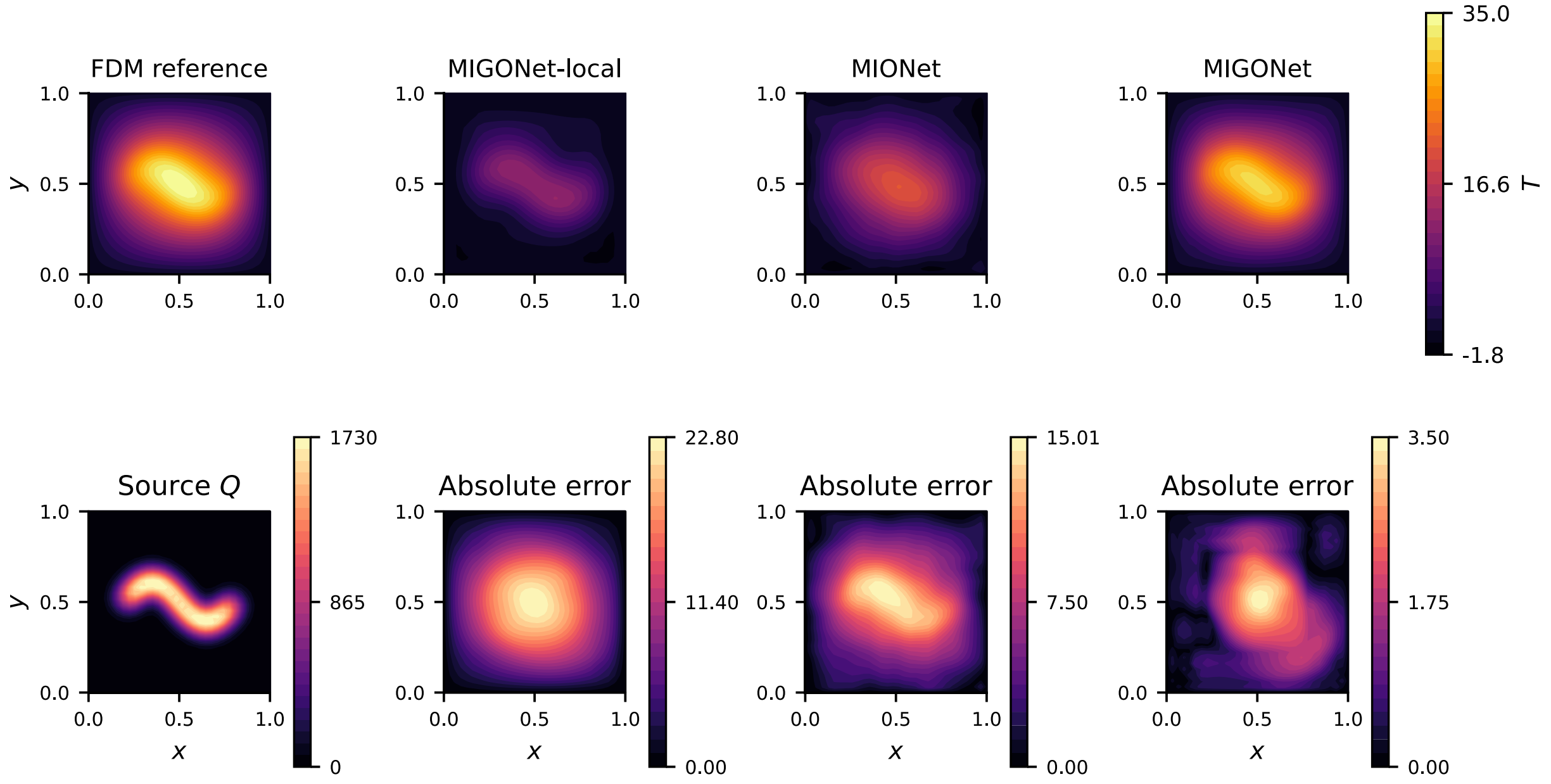

Figure 8. Curved wire heating at $A = 35\,^{\circ}C$ with zero wall temperature. This fixed illustration uses $c = 0.5$, $a = 0.12$, $\phi = 0$, zero rotation, and $\sigma = 0.055$. It is additional to the 36 wire configurations in Table 6.

Figures 8 and 9 show representative wire and Gaussian-spot cases, respectively. Diffusion spreads the temperature response well beyond the region in which heat is generated. MIGONet-local substantially underpredicts this wider response, consistent with the limited distance over which local source information can propagate through six message-passing layers. By supplying a representation of the complete source field to every node before message passing, MIGONet more closely reproduces the nonlocal temperature response, although the remaining error contours show that the approximation is not exact.

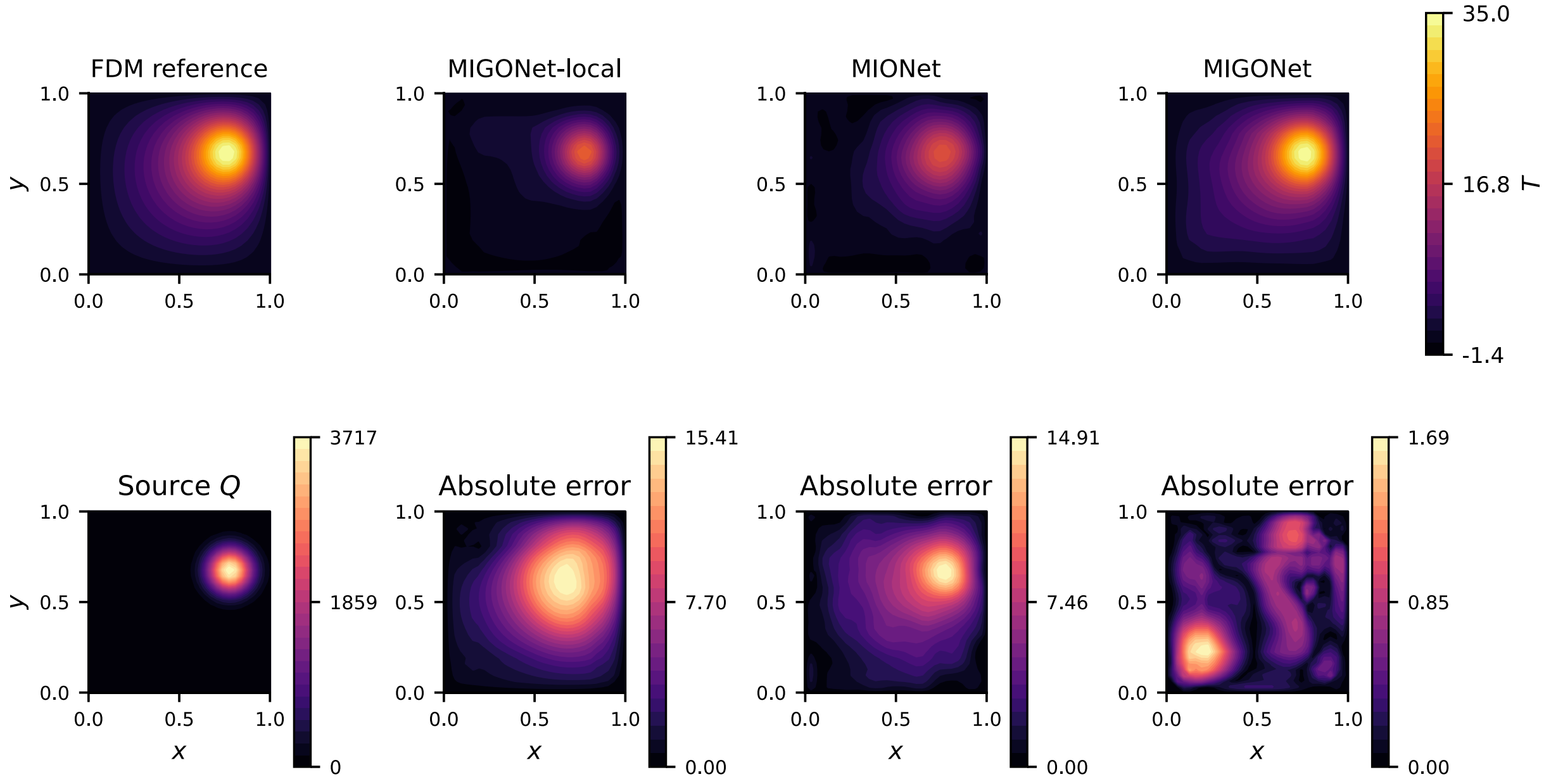

Figure 9. Gaussian spot heating at $A = 35\,^{\circ}C$ with zero wall temperature. The lower row shows the source and absolute errors.

### 6.5 Effect of source magnitude and width

Figure 10 examines sensitivity to source magnitude by evaluating the same wire, Gaussian-spot, and uniform-source configurations at $A = 15$, $35$, and $55\,^\circ\mathrm{C}$. For a fixed source shape with homogeneous Dirichlet boundaries, the discrete diffusion problem is linear and the reference temperature scales with source strength. The comparison therefore indicates how consistently the learned models reproduce this scaling. MIGONet has the lowest average error for every source family at each displayed amplitude. Its mean wire error increases from 8.03% to 16.94%, and its mean spot error increases from 8.36% to 18.12%, showing that the benefit of complete-source information persists across the tested amplitudes even though the approximation error depends on source magnitude.

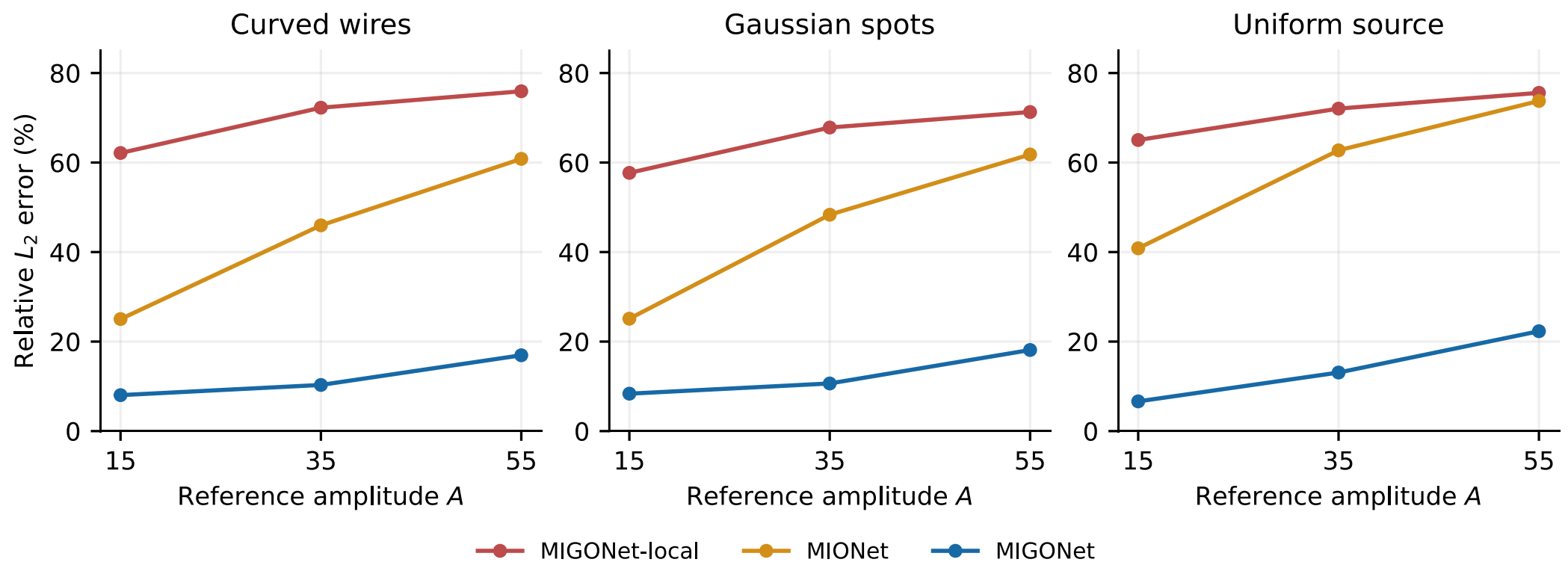

Figure 10. Effect of source strength, expressed through reference peak temperature $A$. Curves average the same source configurations at every amplitude. All displayed amplitudes lie within the training interval.

Figure 11 isolates sensitivity to source localization by holding $A = 35\,^\circ\mathrm{C}$ fixed and varying the wire width. The three widths are selected to span relatively narrow to broader line sources while remaining resolvable on the background grid. MIGONet remains more accurate than both comparison models at all three widths. To verify that the reported trend is not dominated by the reference-grid resolution, the same wire cases are solved on 32 × 32 and 63 × 63 FDM grids; the relative $L_2$ difference at shared nodes remains below 0.20% for every case. This grid comparison supports the reference solutions for the tested finite-width sources.

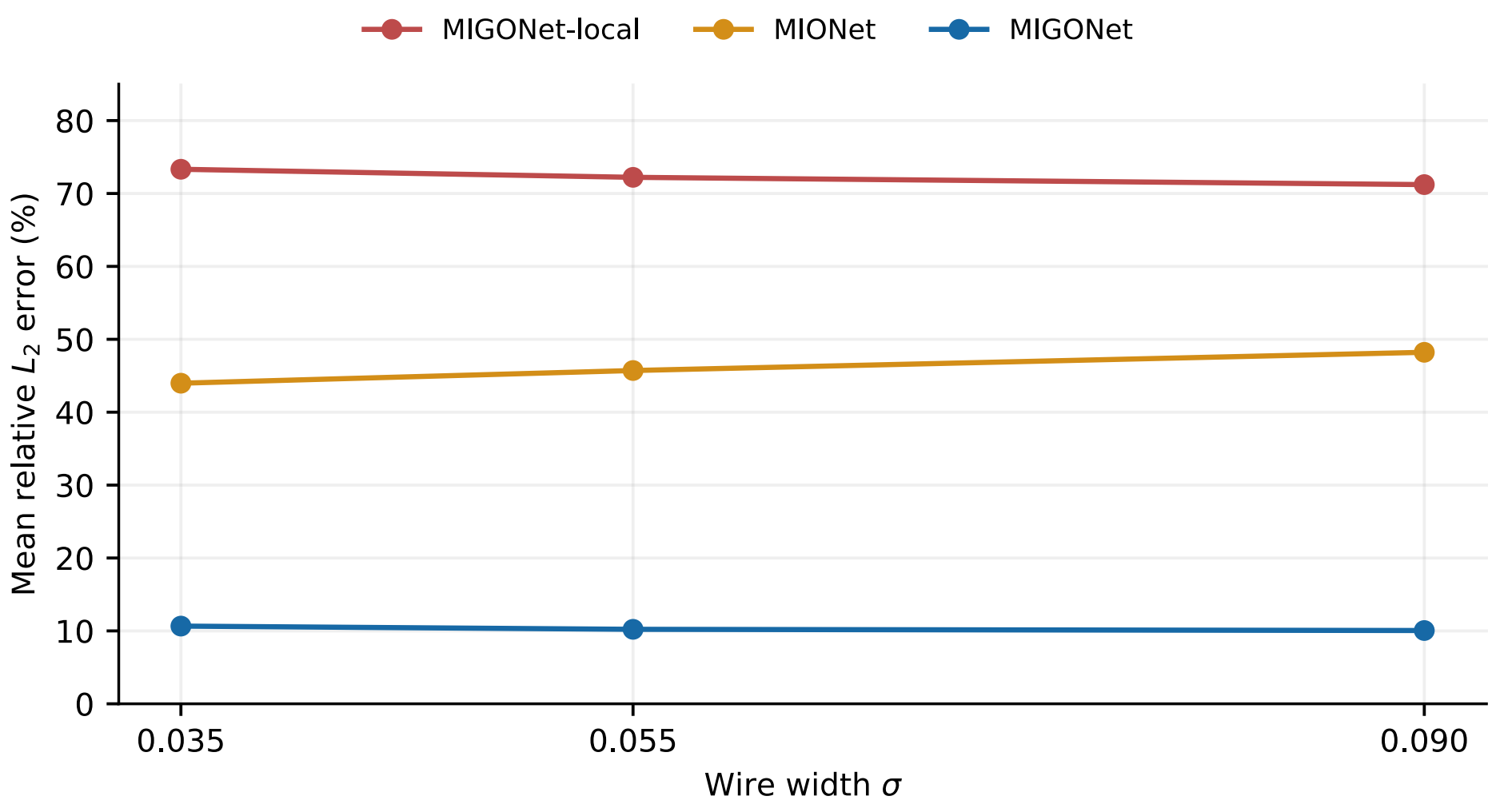

Figure 11. Effect of wire width at $A = 35\,^\circ C$. Each point averages the 12 centerlines at that width.

### 6.6 Computational cost of training-data generation

This section compares the cost of constructing the training data. A conventional supervised pipeline requires a numerical solution for each sampled input, whereas discrete MMS retains the manufactured temperature field directly as the target. All timings were measured on the CPU of a single interactive Texas A&M High Performance Research Computing (HPRC) session [24] (16 threads, Intel Xeon Gold 6248R).

A paired construction is used to isolate the numerical label-solution cost avoided by MMS. For each repetition, the same random manufactured fields are used to construct the source and mixed-boundary inputs. MMS timing includes random-field generation, Gaussian smoothing, amplitude scaling, evaluation of the discrete source, and calculation of the boundary temperatures or fluxes. At that stage, the manufactured temperature field is already available and is retained as the training target. The matched FDM pipeline performs the same input construction and then additionally solves the corresponding discrete system using the iterative formulation in Section 2.3 with the same $10^{-6}$ residual tolerance. Thus, the input-generation work is common to both pipelines, and the additional FDM time measures the label-solution work avoided by discrete MMS. Three independently seeded repetitions are performed for 1,000 and 10,000 samples. Table 7 reports the median wall times; the 10,000-sample case matches the rectangular training-set size used in this study. The data-generation time of the matched FDM pipeline and the resulting speedup are defined as:

$$t_{\text{FDM dataset}} = t_{\text{MMS}} + t_{\text{label solves}}, \qquad S_{\text{data}} = \frac{t_{\text{FDM dataset}}}{t_{\text{MMS}}} \tag{24}$$

where $t_{\text{MMS}}$ is the MMS construction time, $t_{\text{label solves}}$ is the additional time spent solving the discrete system for every sample, and $t_{\text{FDM dataset}}$ is the total time of the matched FDM pipeline. $S_{\text{data}} > 1$ indicates that MMS requires less time than FDM-based labeling.

Table 7. Matched rectangular data-generation times. MMS retains the manufactured temperature directly, whereas the matched FDM pipeline solves the corresponding discrete system using the iterative procedure described in section 2.3.

| **Samples** | **MMS, s** | **FDM pipeline, s** | **Time saved, s** | **Speedup** |
|---|---|---|---|---|
| 1,000 | 0.280 | 48.788 | 48.508 | 174.07× |
| 10,000 | 3.149 | 486.649 | 483.500 | 154.54× |

Table 7 shows that, for 1,000 samples, retaining the manufactured target reduces the measured data-generation time from 48.788 to 0.280 s, corresponding to a 99.43% reduction in wall time and a 174.07× pipeline speedup. At the 10,000-sample training scale, the MMS pipeline requires 3.149 s compared with 486.649 s for the matched FDM pipeline. Discrete MMS therefore avoids approximately 483.5 s of iterative label-solution time, reduces the measured wall time by 99.35%, and gives a 154.54× pipeline speedup. Because both pipelines use the same sampled inputs and the same discrete formulation, this comparison directly quantifies the numerical solution cost avoided when the manufactured field is retained as the training target. Training and inference times are separately reported in Appendix C.

## 7 Application to additional geometries

The additional-geometry studies examine whether the same MMS-based training and MIGONet formulation can be applied to domains with features that change the available heat-flow paths and boundary representation. Three geometries are selected for specific reasons: the L-shape introduces a re-entrant corner and a nonconvex conduction path, the disc introduces a curved external boundary, and the ring introduces an internal boundary and two disconnected boundary components. A separate MIGONet is trained for each geometry using only its own MMS dataset; therefore, these studies assess the applicability of the same data-generation and learning workflow across different domain features rather than transfer of a single trained model between geometries.

All geometries use the same $32 \times 32$ Cartesian background grid. A fixed mask identifies the active nodes of each physical domain, and graph edges connect active horizontal and vertical neighbors. Inactive background nodes are excluded from the normalization statistics, training losses, and evaluation errors, and normalized source values are set to zero outside the active domain. The hidden dimensions, six message-passing layers, global source encoder, node decoder, and optimization settings remain unchanged. Only the active-node mask and boundary-input size change with the geometry, as summarized in Table 8.

Table 8. Geometry representation and model size. Boundary entries include segment endpoints according to the boundary definition.

| Domain | Active nodes | Segments | Boundary inputs | Parameters |
|---|---|---|---|---|
| Rectangle | 1,024 | 4 | 256 | 628,097 |
| L-shape | 768 | 6 | 256 | 628,097 |
| Disc | 740 | 2 | 168 | 616,833 |
| Ring | 552 | 2 | 256 | 628,097 |

The disc and ring are centered at $(0.5,0.5)$. Using the radial coordinate $\rho$, their active-domain masks are defined by Eq. (25).

$$\rho = \sqrt{(x-0.5)^2 + (y-0.5)^2}, \qquad \Omega_{\text{disc}}\colon\ \rho \le 0.5, \qquad \Omega_{\text{ring}}\colon\ 0.25 \le \rho \le 0.5 \tag{25}$$

The continuous definitions in Eq. (25) are represented by the corresponding active grid nodes. Temperature and flux conditions are applied at boundary nodes using the same discrete operators introduced in Section 2 and used for the manufactured data in Section 3. This maintains a common discrete formulation for the training targets and the FDM reference solutions on each geometry.

## 7.1 L-shaped domain

The L-shaped domain is formed by removing the upper-right quadrant of the unit square, producing a re-entrant corner and six boundary segments. The four cases in Figure 12 are selected to examine both zero-source boundary-condition cases and prescribed-source cases in this nonconvex domain at $A = 35\ ^\circ\text{C}$. The first two cases have $Q = 0$ and isolate the effect of the boundary specification: the all-Dirichlet case drives heat diagonally through the two arms, whereas the mixed case replaces the top and bottom outer conditions with Neumann fluxes while retaining Dirichlet conditions on the remaining segments. Their reference fields are defined by Eq. (26), with the maximum evaluated over the active grid.

$$T_{\text{D}} = 35(x - 0.75y), \qquad T_{\text{mixed}} = 35\frac{7x+13y}{\max\limits_{\Omega_h}(7x+13y)} \tag{26}$$

The two prescribed-source cases use zero temperature (homogeneous Dirichlet conditions) on all boundary segments and are chosen to place localized heating in different parts of the domain. The curved wire lies in the lower arm with $c = 0.3$, $a = 0.12$, $\phi = 0$, zero rotation, and $\sigma = 0.055$ in Eq. (23). The Gaussian spot is centered at $(0.25,0.72)$ in the upper arm with width $0.09$. These locations test whether the learned model can represent the diffusion response when heat is introduced on opposite sides of the re-entrant corner. In both cases, the source strengths give a reference peak of $35\ ^\circ\text{C}$.

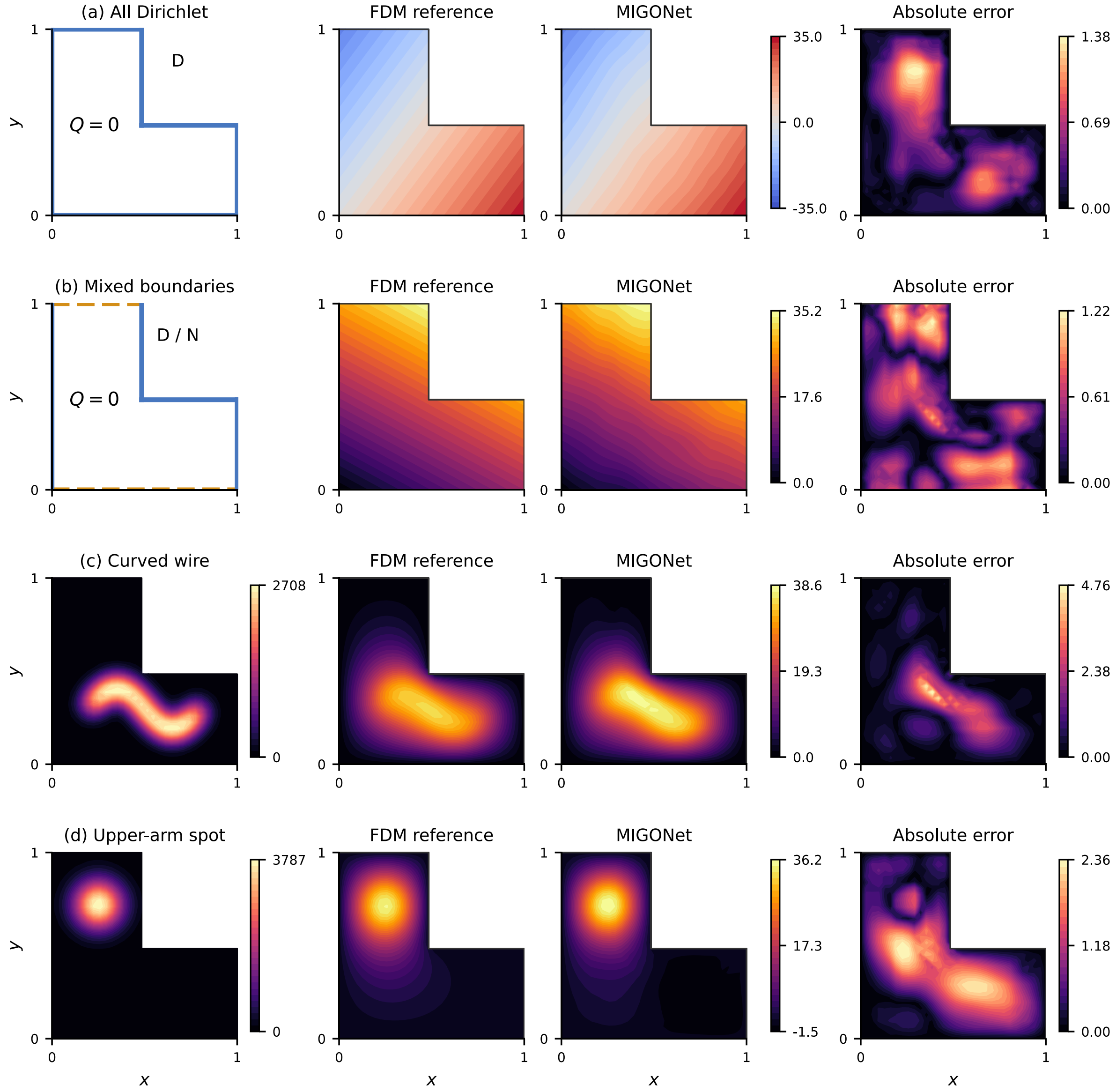

Figure 12. L-shape predictions at $A = 35\,^{\circ}C$. Rows show (a) all-Dirichlet conduction, (b) mixed-boundary conduction, (c) a lower-arm wire, and (d) an upper-arm spot. The first column shows boundary types or source $Q$; solid blue boundaries are Dirichlet and dashed orange boundaries are Neumann.

Table 9. L-shape errors at $A = 35\,^{\circ}C$, evaluated over active nodes.

| **Case** | **Relative $L_2$ error, %** | **Maximum absolute error, $^{\circ}C$** |
|---|---|---|
| All Dirichlet | 3.00 | 1.38 |
| Mixed boundaries | 2.48 | 1.22 |
| Curved wire | 7.01 | 4.76 |
| Upper-arm spot | 10.14 | 2.36 |

The errors for the four L-shaped cases are summarized in Table 9. The all-Dirichlet and mixed-boundary cases give relative errors of 3.00% and 2.48%, respectively, while the curved-wire and upper-arm-spot errors are 7.01% and 10.14%. Figure 12 shows that MIGONet reproduces the principal temperature gradients in the zero-source boundary-condition cases and the broader temperature response generated by the localized heat sources. The prescribed-source cases are more demanding because heat must spread through the nonconvex domain and around the re-entrant corner; the upper-arm spot gives the largest error among the four cases, but the predicted field retains the location and overall extent of the heated region.

## 7.2 Disc-shaped domain

The disc isolates the effect of a curved external boundary while removing the re-entrant corner present in the L-shape. Figure 13 considers two zero-source boundary-condition cases and two prescribed-source cases at $A = 35\ ^\circ\mathrm{C}$. The first two cases use $Q = 0$. The all-Dirichlet case prescribes temperature on both boundary arcs, whereas the mixed case prescribes temperature on the upper arc and outward flux on the lower arc. Their reference temperatures follow $T = AF/\max_{\Omega_h}|F|$, using $F = (x - 0.5) + 0.5(y - 0.5)$ for the all-Dirichlet case and $F = 0.2 + x + 0.6y$ for the mixed case. These linear fields are selected to evaluate the curved-boundary and normal-flux treatment without introducing internal heat generation.

The prescribed-source cases use zero temperature on both boundary arcs and are selected to compare two distinct radial heating patterns: a compact Gaussian spot at the center and a circular heating band distributed around the interior of the disc. These cases test centrally concentrated heating against heating distributed along a curved path. Their radial source functions are:

$$Q_{\text{spot}} = \alpha \exp\left(-\frac{\rho^2}{2\sigma^2}\right), \qquad Q_{\text{band}} = \alpha \exp\left[-\frac{(\rho-\rho_s)^2}{2\sigma^2}\right] \tag{27}$$

The central spot uses $\sigma = 0.12$, while the circular band uses radius $\rho_s = 0.25$ and width $\sigma = 0.07$. As in the rectangular study, the source magnitude $\alpha$ is scaled so that the FDM reference peak is $A = 35\ ^\circ\mathrm{C}$. The network receives only the source and boundary data; the target temperature is not supplied as an input.

Table 10 shows relative errors of 1.52% for the all-Dirichlet case, 1.18% for the mixed-boundary case, 1.25% for the central spot, and 1.26% for the circular heating band. The corresponding fields in Figure 13 show close agreement between the FDM references and MIGONet predictions for both zero-source boundary-condition cases and prescribed-source cases. In particular, the central spot produces a localized peak near the center, whereas the circular band produces a broader warm region inside the heated path; MIGONet reproduces both radial responses and their decay toward the cooled boundary. The similar errors across the four cases indicate that the curved outer boundary does not introduce a marked loss of accuracy for these evaluations.

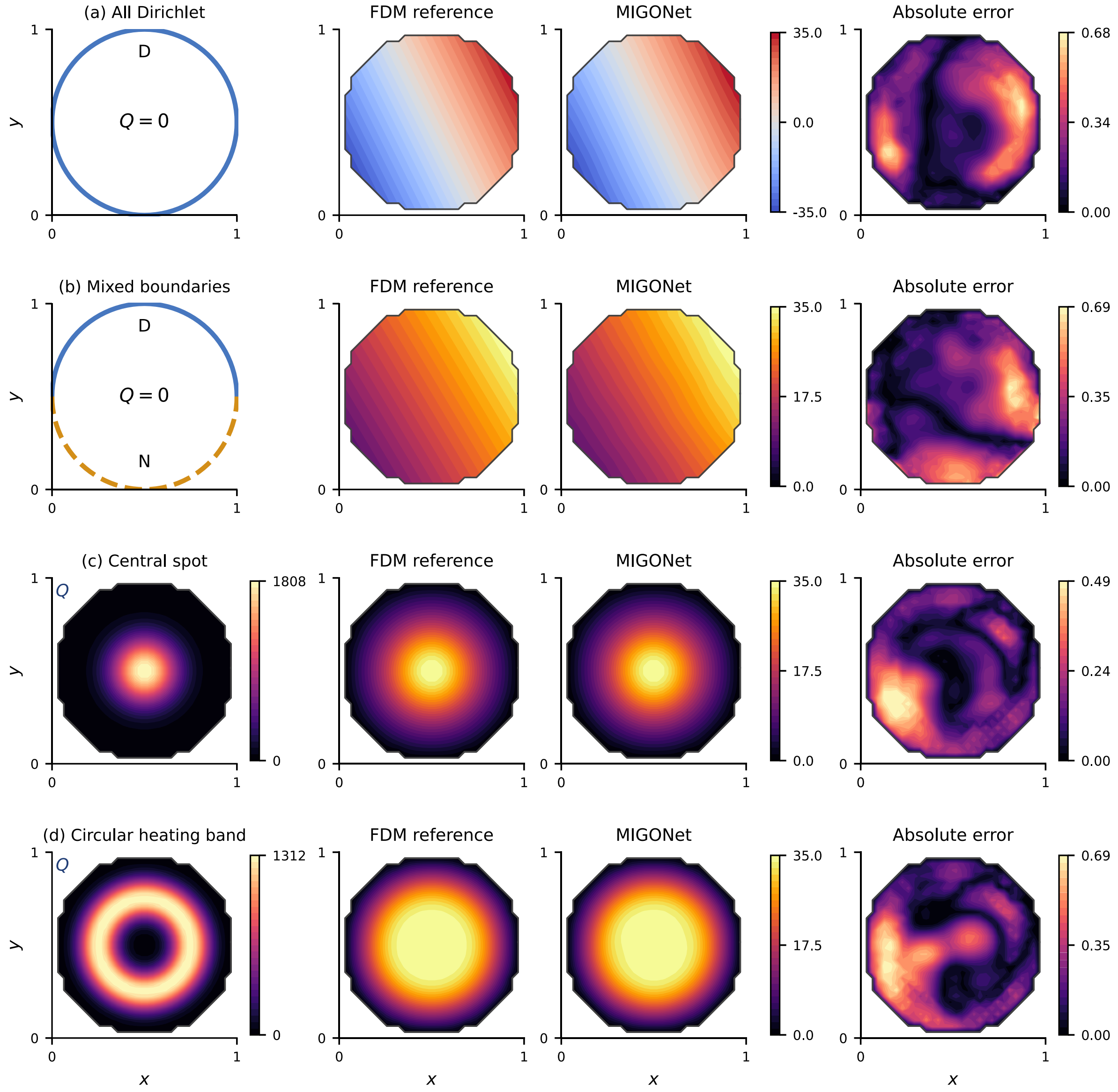

Figure 13. Disc predictions at $A = 35\,^{\circ}C$. Rows show (a) all-Dirichlet conduction, (b) mixed-boundary conduction, (c) a central spot, and (d) a circular heating band. Solid blue arcs denote Dirichlet conditions and dashed orange arcs denote Neumann conditions. The first column shows boundary conditions or source $Q$.

Table 10. Disc errors at $A = 35\,^{\circ}C$, evaluated over active nodes.

| **Case** | **Relative $L_2$ error, %** | **Maximum absolute error, $^{\circ}C$** |
|---|---|---|
| All Dirichlet | 1.52 | 0.68 |
| Mixed boundaries | 1.18 | 0.69 |
| Central spot | 1.25 | 0.49 |
| Circular heating band | 1.26 | 0.69 |

## 7.3 Ring-shaped domain

The ring adds an internal boundary and therefore provides the most distinct boundary topology among the three additional geometries. The four cases in Figure 14 are selected to test heat transfer across the annulus, mixed conditions on the inner and outer boundaries, and heating distributed over part or all of the circumference. In the first boundary-condition case, the inner boundary is held at $T = 35\,°C$, the outer boundary at $T = 0$, and $Q = 0$, producing conduction from the inner surface through the annular region. The mixed case prescribes temperature on the outer boundary and flux on the inner boundary, with reference field $T = 35F/\max_{\Omega_h}|F|$ and $F = 1 + 0.4(x - 0.5) + 0.7(y - 0.5)$. This case specifically exercises the inner-boundary normal convention introduced in Section 2, for which the outward normal of the physical domain points toward the hole.

The prescribed-source cases use zero temperature on both boundaries and place the heating approximately midway between the inner and outer surfaces. The arc heater uses the finite-width line-source construction of Eq. (22), with $\sigma = 0.055$ and centerline.

$$\mathbf{r}_w(\theta) = (0.5 + 0.375\cos\theta,\ 0.5 + 0.375\sin\theta), \qquad \pi/6 \le \theta \le 5\pi/6 \tag{28}$$

The annular heating band uses Eq. (27) with $\rho_s = 0.375$ and $\sigma = 0.055$. The radius places both sources midway between the inner and outer boundaries. The arc case tests localized circumferential heating over part of the ring, whereas the annular band tests heating around the full circumference. Their source strengths are scaled to a reference peak of $A = 35\,°\mathrm{C}$.

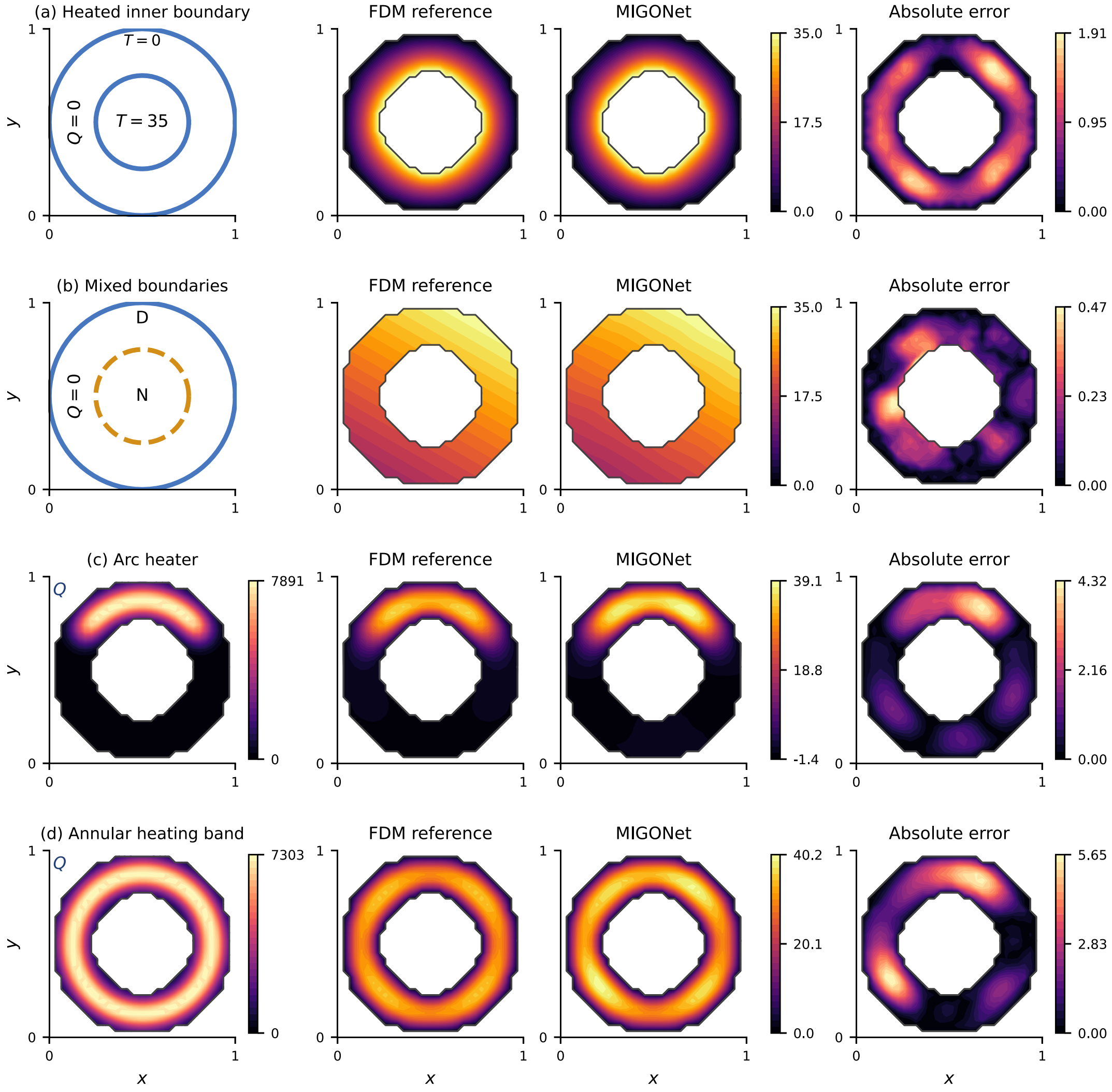

Figure 14. Ring predictions at $A = 35\,^{\circ}C$. Rows show (a) a heated inner boundary and cooled outer boundary, (b) an outer Dirichlet boundary and inner Neumann boundary, (c) an arc heater, and (d) an annular heating band.

Table 11. Ring errors at $A = 35\,^{\circ}C$, evaluated over active nodes.

| **Case** | **Relative $L_2$ error, %** | **Maximum absolute error, $^{\circ}C$** |
|---|---|---|
| Heated inner boundary | 4.61 | 1.91 |
| Mixed boundaries | 0.59 | 0.47 |
| Arc heater | 9.91 | 4.32 |
| Annular heating band | 9.34 | 5.65 |

The ring results are summarized in Table 11. The heated-inner-boundary and mixed-boundary cases give relative errors of 4.61% and 0.59%, respectively, while the arc-heater and annular-band errors are 9.91% and 9.34%. Figure 14 shows that MIGONet captures the radial heat transfer between the two boundary components and places the warm regions for prescribed-source cases along the intended heated paths. The larger errors for the prescribed-source cases are consistent with the additional spatial structure introduced by the circumferential heating distributions, while both remain below 10% for the cases shown.

Taken together, the twelve main cases span zero-source boundary-condition cases and prescribed-source cases in domains with a re-entrant corner, a curved outer boundary, and an internal hole. Relative errors range from 0.59% to 10.14%. Because each geometry is trained separately, these results do not represent geometric extrapolation by a single network; instead, they show that the same discrete MMS data-generation procedure and MIGONet architecture can be applied without changing the core learning formulation across these distinct domain features. Appendix A evaluates the same main cases at $A = 15$, $35$, and $55\ ^{\circ}\mathrm{C}$, and reports additional source locations and uniform-heating cases to examine behavior beyond the single amplitude illustrated in Figures 12-14. Because each geometry uses its own training dataset, trained model, and evaluation cases, the results demonstrate the applicability of the same workflow across varying domains and testing cases, rather than a controlled comparison between geometries.

## 8 Summary

This study combines discrete manufactured-solution data generation with a multiple-input graph neural operator, using steady heat conduction as the model problem. The MMS procedure constructs mutually consistent source, boundary, and temperature data using the same discrete operators employed by the numerical formulation, while avoiding a PDE solution for every training sample. MIGONet combines local node features, global boundary information, and an embedding of the complete source field, allowing every node to receive information about heat generation throughout the domain before local message passing. Dirichlet conditions prescribe known boundary temperatures and are enforced exactly, whereas Neumann conditions prescribe heat flux, for which the boundary temperature depends on the interior solution. Neumann conditions are therefore supplied as network inputs and learned from the training data, and the predictions do not satisfy exact discrete flux conservation.

At $A = 35\ ^{\circ}C$ in the rectangle, MIGONet gives mean relative $L_2$ errors of 10.31% for curved wires, 10.64% for Gaussian spots, and 13.08% for uniform heating. The corresponding MIGONet-local errors are 72.25%, 67.82%, and 72.05%, while MIONet also gives substantially larger errors for these source families. These comparisons support global source encoding for diffusion problems in which localized heat generation produces a temperature response extending well beyond the local message-passing neighborhood.

The L-shape, disc, and ring studies use the same MMS-only training procedure with geometry-specific masks and boundary inputs. Across the twelve main additional-geometry cases, relative errors range from 0.59% to 10.14%. The resulting models represent heat conduction around a re-entrant corner, along a curved boundary, and across an annular region, demonstrating that the same data-generation and learning workflow can be applied to distinct domain features while retaining mixed-boundary and prescribed-source flexibility.

The data-generation comparison shows a clear computational benefit of discrete MMS over conventional FDM-based labeling. For 10,000 matched samples, discrete MMS requires 3.149 s compared with 486.649 s when the same source and boundary inputs are independently labeled by FDM, reducing the measured wall time by 99.35% and giving a 154.54× pipeline speedup. The paired comparison therefore directly measures the label-solution cost avoided by retaining the manufactured field.

The present results are limited to two-dimensional, constant-conductivity diffusion on fixed Cartesian discretizations, with a separate model trained for each geometry. The discrete MMS procedure only requires the discrete operator to be evaluated on a prescribed field and therefore extends directly to variable conductivity, advection–diffusion, and transient problems through the corresponding discrete flux, convective, and accumulation terms. These extensions require additional operator inputs, namely the conductivity field, the velocity field, or the previous temperature field and time step, and remain to be demonstrated. The learned predictions also retain amplitude-dependent errors, and Neumann conditions are represented through the inputs rather than enforced exactly. Agreement with the discrete temperature references therefore does not by itself establish accurate boundary fluxes, global heat conservation, or transfer across geometries and resolutions. Taken together, these results demonstrate that discrete MMS and global source-aware graph operator learning can reduce the computational cost of constructing consistent training datasets while retaining flexibility across source distributions, mixed boundary conditions, and domain geometries.

## Appendix A Additional geometry results

Figure A.1 evaluates the four main cases for each geometry at $A = 15, 35$, and $55\ {}^\circ C$. The boundary pattern or source shape stays fixed while its magnitude changes. All three amplitudes lie within the training interval. This comparison shows the amplitude response for the same physical configurations, without changing the geometry or retraining the model.

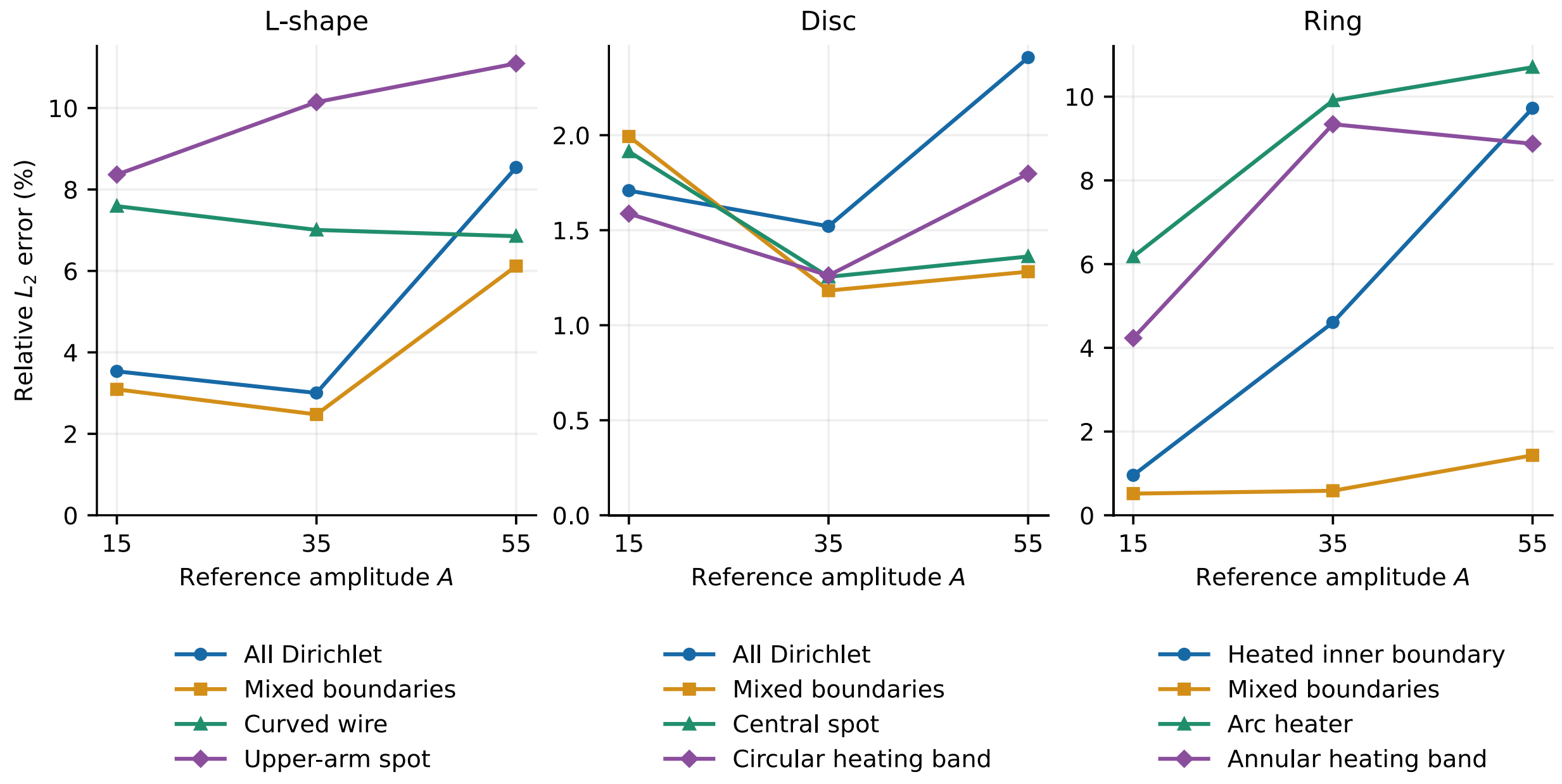

Figure A.1. Amplitude response for the main L-shape, disc, and ring cases. Each line follows one fixed configuration. Errors use active nodes only. The maximum absolute reference temperature is $A$ for zero-source boundary-condition cases, and the peak reference temperature is $A$ for the positive-source cases.

Table A.1 reports additional configurations at $A = 35\ {}^\circ C$. For the L-shape, the lower-arm spot complements the upper-arm spot in the main text. The disc spot is moved away from the center, and the ring spot concentrates heating over a shorter part of the circumference than the arc heater. Uniform-source cases provide a distributed-heating comparison. The L-shape curved-contour case uses $T = 35(x^2 - y^2)$, with temperatures applied on all boundaries and $Q = 0$.

Table A.1. Additional geometry cases at $A = 35\,^{\circ}C$.

| **Domain** | **Case** | **Relative $L_2$ error, %** | **Maximum absolute error, $^{\circ}C$** |
|---|---|---|---|
| L-shape | Curved contours | 4.06 | 1.92 |
| L-shape | Lower-arm spot | 12.44 | 4.39 |
| L-shape | Uniform heating | 6.51 | 5.00 |
| Disc | Offset spot | 1.59 | 0.58 |
| Disc | Uniform heating | 2.04 | 0.79 |
| Ring | Localized spot | 11.11 | 2.12 |
| Ring | Uniform heating | 12.43 | 6.57 |

Figure A.2 shows the three additional spot cases. The Gaussian centers and widths are $(0.72, 0.25)$ and $\sigma = 0.09$ in the L-shape, $(0.68, 0.5)$ and $\sigma = 0.10$ in the disc, and $(0.5, 0.875)$ and $\sigma = 0.055$ in the ring. All boundary temperatures are zero, and source strengths are scaled to a reference peak of $35\,^{\circ}\mathrm{C}$.

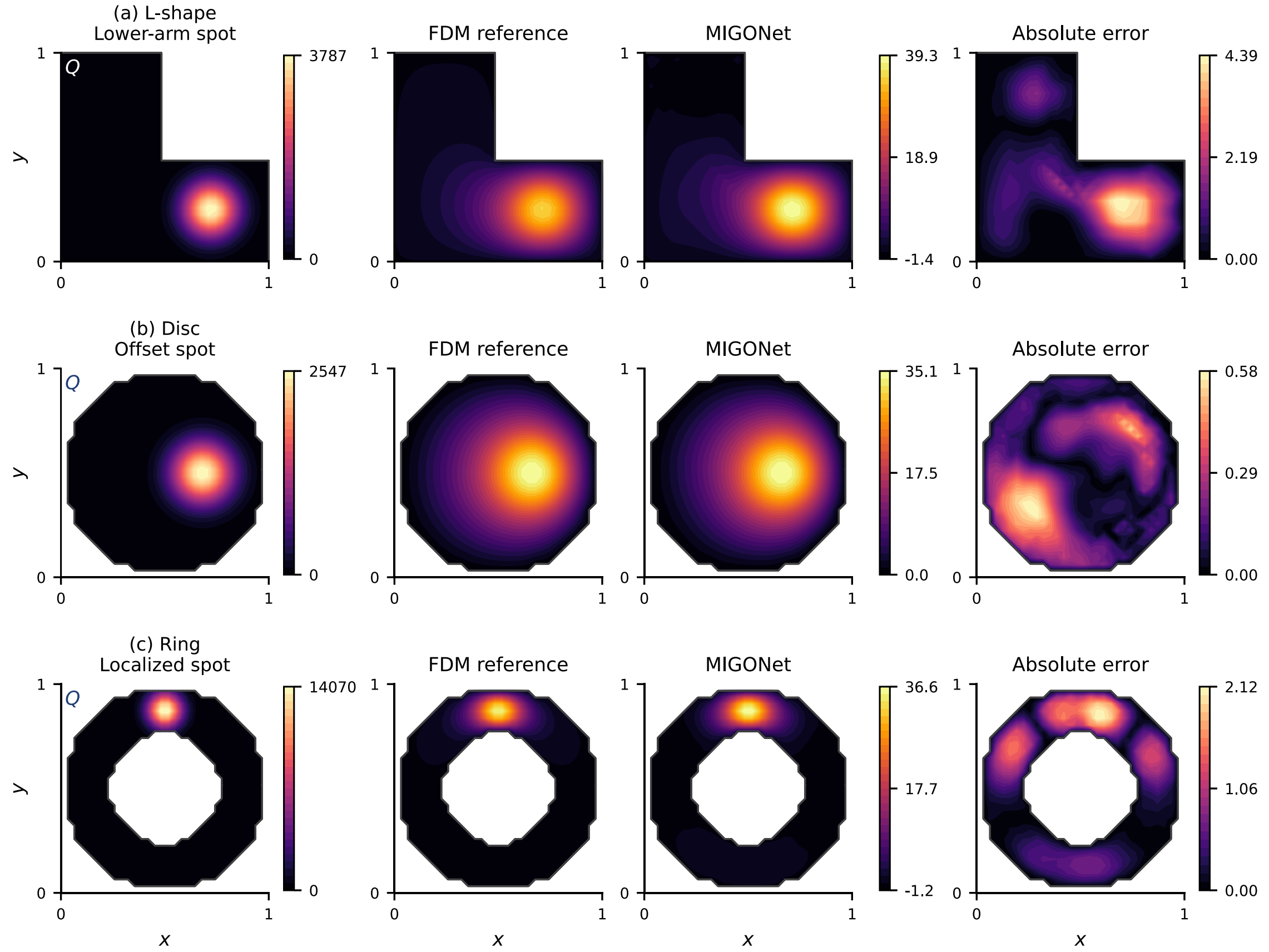

Figure A.2. Additional localized heating at $A = 35\,^{\circ}C$: (a) lower-arm spot in the L-shape, (b) offset spot in the disc, and (c) localized spot in the ring. Columns show the source location and distribution, FDM reference, MIGONet prediction, and absolute error.

## Appendix B Localized-source implementation details

This appendix provides the numerical implementation details for the curved-wire, Gaussian-spot, and uniform-heating evaluation cases introduced in Section 6.4. The construction is separated from the main results discussion to keep Section 6.4 focused on the physical interpretation and model comparison while documenting the source generation, quadrature, parameter sampling, amplitude scaling, and grid-resolution check used in the evaluation.

### B.1 Curved-wire centerline and numerical line integral

The unrotated wire centerline is parameterized by $s \in [0,1]$ as given in Eq. (23) of the main text. In the implementation, the curve is sampled at $M = 161$ uniformly spaced parameter values, $s_j = \frac{j}{160}$ for $j = 0, 1, \ldots, 160$. The corresponding unrotated position vector is $r_0(s) = (0.2 + 0.6s,\ c + a\,sin(2\pi s + \varphi))$.

The sampled centerline is rotated about the center of the unit square, $(0.5,0.5)$, by an orientation angle $\theta$. For an unrotated point $(x_0, y_0)$, the rotated coordinates are:

$$x_r = 0.5 + cos(\theta)(x_0 - 0.5) - sin(\theta)(y_0 - 0.5),$$
$$y_r = 0.5 + sin(\theta)(x_0 - 0.5) + cos(\theta)(y_0 - 0.5). \quad \text{(B1)}$$

Rotation changes the wire position and orientation but preserves the centerline arclength. The continuous line integral in Eq. (22) is evaluated numerically using the sampled centerline. With $\Delta s = 1/160$, the local arclength weight is approximated as:

$$\Delta\ell_j \approx \|\frac{dr_0}{ds}(s_j)\|\,\Delta s, \quad \Delta s = \frac{1}{160}. \quad \text{(B2)}$$

The centerline derivatives are evaluated numerically from the sampled coordinates. The first and last arclength weights are multiplied by one half, giving the composite trapezoidal treatment used in the implementation. At each Cartesian grid node $x_i$, the unscaled wire source is evaluated as:

$$Q_\sigma^{(0)}(x_i) \approx \frac{1}{\sqrt{2\pi}\sigma} \sum_{j=0}^{160} exp[-\frac{\|x_i - r(s_j)\|^2}{2\sigma^2}]\,\Delta\ell_j. \quad \text{(B3)}$$

Equation (B3) is the discrete counterpart of the line integral in Eq. (22). Each sampled centerline segment contributes according to its distance from the grid node: nearby segments contribute strongly, whereas distant segments contribute negligibly. The factor $\frac{1}{\sqrt{2\pi}\sigma}$ provides the Gaussian normalization across the wire width. The superscript $(0)$ denotes the unscaled source before the reference-temperature amplitude is imposed.

### B.2 Wire-parameter sampling

A fixed random-number seed of 91002 is used to construct the rectangular evaluation sources. Twelve base wire centerlines are generated by independently sampling:

$$c \sim U(0.42,0.58), \quad a \sim U(0.02,0.14), \quad \varphi \sim U(0,2\pi), \quad \theta \sim U(-0.5,0.5)\,rad. \quad \text{(B4)}$$

Each centerline is evaluated at three source widths, $\sigma = 0.035, 0.055, 0.09$, yielding 36 wire cases at each reported amplitude. The fixed wire shown in Figure 8 is used only for visualization and is additional to these 36 cases; it uses $c = 0.5,\ a = 0.12,\ \varphi = 0,\ \theta = 0,\ \sigma = 0.055$.

### B.3 Gaussian spots and uniform heating

The same deterministic evaluation sequence generates twelve Gaussian spots. Their center coordinates are sampled independently as $c_x, c_y \sim U(0.2,0.8)$, and the spot width is sampled as $\sigma_{spot} \sim U(0.07,0.18)$. The unscaled spot source is:

$$Q_{spot}^{(0)}(x,y) = exp[-\frac{(x-c_x)^2+(y-c_y)^2}{2\sigma_{spot}{}^2}]. \quad \text{(B5)}$$

The uniform-heating case uses a constant unscaled source over the complete rectangular domain. All wire, spot, and uniform-source cases use homogeneous Dirichlet temperature, $T = 0$, on all four walls.

### B.4 Reference-amplitude scaling

For each positive source case, the source magnitude is set through the reference peak temperature rather than by prescribing a fixed source amplitude. Let $Q^{(0)}$ denote the unscaled source field and $T^{(0)}$ the FDM solution obtained from $Q^{(0)}$ with zero temperature on all walls. The scaling factor and scaled fields are:

$$\alpha = \frac{A}{max_{\Omega}\, T^{(0)}}, \quad Q = \alpha Q^{(0)}, \quad T = \alpha T^{(0)}. \quad \text{(B6)}$$

Because the steady diffusion problem is linear for the fixed conductivity and boundary conditions used here, this scaling gives a reference solution with peak temperature $A$ without changing the spatial source shape. The rectangular comparisons use $A = 15, 35, 55\ °C$. The network receives the scaled source and boundary data but is not given the target temperature or target peak as an input.

### B.5 Grid-resolution check for the wire sources

To assess whether the finite wire widths are adequately represented on the $32 \times 32$ evaluation grid, each of the 36 wire configurations is also solved on a $63 \times 63$ FDM grid using the same continuous centerline parameters and source width. Because both grids span the same unit square, every second node of the $63 \times 63$ grid coincides with a node of the $32 \times 32$ grid. The relative difference at the shared nodes is evaluated as:

$$\varepsilon_{grid} = 100\ \frac{\|T_{32} - T_{63,shared}\|_2}{\|T_{63,shared}\|_2}. \quad \text{(B7)}$$

Across the 36 wire cases, the relative difference ranges from approximately 0.095% to 0.197% and remains below 0.20% for every tested width and centerline. This check supports the use of the $32 \times 32$ reference solutions for the finite-width sources considered in the manuscript; it is not a convergence study for an ideal zero-width line source.

## Appendix C Training and inference times

The rectangular MIGONet was trained using the settings in Table 2 and stopped early after approximately 290 epochs upon convergence. Training required approximately 36 min, with a median epoch time of 7.1 s. Table C.1 compares the prediction time of the trained model with the iterative FDM solution time for representative cases. It can be noticed from Table C.1 that the prediction time of MIGONet is nearly constant across the cases, whereas the iterative FDM solution time increases for mixed boundary conditions. The largest speedup, 29.69×, is obtained for the applied wall flux case.

Table C.1. Execution times for rectangular MIGONet prediction and iterative FDM solution at A = 35 °C. The iterative solver uses CG for all-Dirichlet systems and GMRES for mixed Dirichlet–Neumann systems, with a relative algebraic residual tolerance of $10^{-6}$.

| Individual case | MIGONet (ms) | Iterative FDM (ms) | Speedup |
|---|---|---|---|
| All-Dirichlet walls | 1.176 | 1.639 | 1.39× |
| Insulated side walls | 1.182 | 15.345 | 12.98× |
| Applied wall flux | 1.179 | 35.004 | 29.69× |
| Curved wire | 1.048 | 1.715 | 1.64× |
| Gaussian spot | 1.152 | 2.071 | 1.80× |
| Uniform heating | 1.166 | 1.520 | 1.30× |

**Author contributions:** All authors have accepted responsibility for the entire content of this manuscript and approved its submission.

**Acknowledgements:** The authors gratefully acknowledge the support from Texas A&M University High Performance Research Computing (HPRC), the James J. Cain '51 Professor III, and the Marcus C. Easterling Chair Funds.

**Data Availability:** The data that support the findings of this study are available from the corresponding author upon reasonable request.

**Research ethics:** Not applicable.

**Conflict of interest:** The authors state no conflict of interest.

**Use of Large Language Models, AI and Machine Learning Tools:** During the preparation of this manuscript, the authors used ChatGPT and Claude to assist with language editing, code troubleshooting, and the post-processing and visualization of results. All resulting code, figures, and manuscript content were reviewed and verified by the authors, who take full responsibility for the analyses, results, and conclusions.

**Research funding:** None declared.

## References

[1] O. Ghattas and K. Willcox, "Learning physics-based models from data: perspectives from inverse problems and model reduction," *Acta Numerica*, vol. 30, pp. 445–554, May 2021, doi: 10.1017/S0962492921000064.

[2] H. Viswanath *et al.*, "Neural Operator: Is Data All You Need to Model the World? An Insight Into the Paradigm of Data-Driven Scientific ML," *IEEE Trans. Pattern Anal. Mach. Intell.*, vol. 48, no. 8, Aug. 2026, doi: 10.1109/TPAMI.2026.3682604.

[3] L. Lu *et al.*, "A comprehensive and fair comparison of two neural operators (with practical extensions) based on FAIR data," *Comput. Methods Appl. Mech. Eng.*, vol. 393, p. 114778, Apr. 2022, doi: 10.1016/J.CMA.2022.114778.

[4] A. Sojitra and O. San, "Method of Manufactured Learning for Solver-free Training of Neural Operators," Nov. 2025, doi: 10.48550/arXiv.2511.12890.

[5] N. Kovachki *et al.*, “Neural Operator: Learning Maps Between Function Spaces With Applications to PDEs,” *Journal of Machine Learning Research*, vol. 24, no. 89, pp. 1–97, 2023, Accessed: Jun. 28, 2026. [Online]. Available: http://jmlr.org/papers/v24/21-1524.html

[6] L. Lu, P. Jin, G. Pang, Z. Zhang, and G. E. Karniadakis, “Learning nonlinear operators via DeepONet based on the universal approximation theorem of operators,” *Nat. Mach. Intell.*, vol. 3, no. 3, pp. 218–229, Mar. 2021, doi: 10.1038/S42256-021-00302-5.

[7] Z. Li *et al.*, “Fourier Neural Operator for Parametric Partial Differential Equations,” *ICLR 2021 - 9th International Conference on Learning Representations*, Oct. 2020, doi: 10.48550/arXiv.2010.08895.

[8] P. Jin, S. Meng, and L. Lu, “MIONet: Learning Multiple-Input Operators via Tensor Product,” *SIAM Journal on Scientific Computing*, vol. 44, no. 6, pp. A3490–A3514, Nov. 2022, doi: 10.1137/22M1477751.

[9] Z. Li *et al.*, “Neural Operator: Graph Kernel Network for Partial Differential Equations,” Mar. 2020, doi: 10.48550/arXiv.2003.03485.

[10] M. Hasnain, N. K. Anand, A. Srinivasa, and P. Kumar, “A novel residual multilayer perceptron-based graph convolutional network for modeling heat transfer and fluid flow,” *Physics of Fluids*, vol. 37, no. 10, p. 107138, Oct. 2025, doi: 10.1063/5.0292956.

[11] M. Hasnain, N. K. Anand, P. Kumar, A. Srinivasa, and S. Ji, “A Physics-Regularized Graph Neural Surrogate for Geometry-Generalizable Fluid Flow Prediction,” 2026, doi: 10.2139/SSRN.6446623.

[12] T. Pfaff, M. Fortunato, A. Sanchez-Gonzalez, and P. W. Battaglia, “Learning Mesh-Based Simulation with Graph Networks,” in *International Conference on Learning Representations (ICLR)*, 2021. Accessed: Aug. 28, 2025. [Online]. Available: https://doi.org/10.48550/arXiv.2010.03409

[13] U. Alon and E. Yahav, “On the Bottleneck of Graph Neural Networks and its Practical Implications,” *ICLR 2021 - 9th International Conference on Learning Representations*, Jun. 2021, doi: 10.48550/arXiv.2006.05205.

[14] S. Wang, H. Wang, and P. Perdikaris, “Learning the solution operator of parametric partial differential equations with physics-informed DeepONets,” *Sci. Adv.*, vol. 7, no. 40, pp. 8605–8634, Oct. 2021, doi: 10.1126/sciadv.abi8605.

[15] R. Bischof, M. Piovarči, M. A. Kraus, S. Mishra, and B. Bickel, “HyPINO: Multi-Physics Neural Operators via HyperPINNs and the Method of Manufactured Solutions,” in *Advances in Neural Information Processing Systems 38 (NeurIPS 2025)*, 2025. doi: 10.48550/arXiv.2509.05117.

[16] W. You, S. Zhou, and X. Meng, “Self-supervised neural operator for solving partial differential equations,” Sep. 2025, doi: 10.48550/arXiv.2509.00867.

[17] Z. Liu, H. Wang, H. Zhang, K. Bao, X. Qian, and S. Song, “Render unto Numerics: Orthogonal Polynomial Neural Operator for PDEs with Non-periodic Boundary Conditions,” *SIAM Journal on Scientific Computing*, vol. 46, no. 4, pp. C323–C348, Jun. 2024, doi: 10.1137/23M1556320.

[18] N. Saad, G. Gupta, S. Alizadeh, and D. C. Maddix, “Guiding continuous operator learning through Physics-based boundary constraints,” *11th International Conference on Learning Representations, ICLR 2023*, Dec. 2022, doi: 10.48550/arXiv.2212.07477.

[19] X. Zhang and K. Garikipati, “Label-free learning of elliptic partial differential equation solvers with generalizability across boundary value problems,” *Comput. Methods Appl. Mech. Eng.*, vol. 417, Dec. 2023, doi: 10.1016/J.CMA.2023.116214.

[20] P. J. Roache, “Code Verification by the Method of Manufactured Solutions,” *J. Fluids Eng.*, vol. 124, no. 1, pp. 4–10, Mar. 2002, doi: 10.1115/1.1436090.

[21] E. Hasani and R. A. Ward, “Generating synthetic data for neural operators,” *SMAI Journal of Computational Mathematics*, vol. 11, pp. 497–516, 2025, doi: 10.5802/smai-jcm.132.

[22] H. Dong, H. Wang, H. Liu, J. Luo, and J. Wang, “Accelerating PDE Data Generation via Differential Operator Action in Solution Space,” *International Conference on Machine Learning (ICML), Proceedings of Machine Learning Research (PMLR)*, vol. 235, pp. 11395–11411, May 2024, doi: 10.48550/arXiv.2402.05957.

[23] S. Choubey, S. Rahi, B. Pal, and M. Agrawal, “A novel data generation scheme for surrogate modelling with deep operator networks,” *Eng. Appl. Artif. Intell.*, vol. 161, Dec. 2025, doi: 10.1016/j.engappai.2025.112086.

[24] “High Performance Research Computing.” Accessed: Sep. 10, 2025. [Online]. Available: https://hprc.tamu.edu/